\documentclass[preprint,12pt]{elsarticle}
\usepackage{natbib}
\usepackage{epsfig}  
\usepackage{subfigure}
\usepackage{algorithmic}
\usepackage{algorithm}
\usepackage{textcomp}
\usepackage{multirow}
\usepackage{verbatim}
\usepackage{amsmath}
\usepackage{mathrsfs}
\usepackage{amsfonts}
\usepackage{amssymb}
\usepackage{amsthm}
\usepackage{subfigure}
\usepackage{sectsty}
\DeclareMathOperator*{\nut}{arg}
\newtheorem{theorem}{Theorem}

\newtheorem{proposition}{Proposition}

\newtheorem{assumption}{Assumption}
\newtheorem{claim}{Claim}

\journal{Pattern Recognition}

\begin{document}

\begin{frontmatter}



\title{Clustering and Latent Semantic Indexing\\Aspects of the Singular Value Decomposition}


\author{Andri Mirzal}
\ead{andrimirzal@utm.my}
\address{Faculty of Computer Science and Information Systems N28-438-20, \\
Universiti Teknologi Malaysia,\\ 
81310 UTM Johor Bahru, Malaysia\\
Phone: +60-755-32260}

\begin{abstract}
This paper discusses clustering and latent semantic indexing (LSI) aspects of the singular value decomposition (SVD). The purpose of this paper is twofold. The first is to give an explanation on how and why the singular vectors can be used in clustering. And the second is to show that the two seemingly unrelated SVD aspects actually originate from the same source: related vertices tend to be more clustered in the graph representation of lower rank approximate matrix using the SVD than in the original semantic graph. Accordingly, the SVD can improve retrieval performance of an information retrieval system since queries made to the approximate matrix can retrieve more relevant documents and filter out more irrelevant documents than the same queries made to the original matrix. By utilizing this fact, we will devise an LSI algorithm that mimicks SVD capability in clustering related vertices. Convergence analysis shows that the algorithm is convergent and produces a unique solution for each input. Experimental results using some standard datasets in LSI research show that retrieval performances of the algorithm are comparable to the SVD's. In addition, the algorithm is more practical and easier to use because there is no need to determine decomposition rank which is crucial in driving retrieval performance of the SVD.
\end{abstract}


\end{frontmatter}


\section{Introduction}\label{ch2:introduction}
There are some literatures that have discussed clustering and latent semantic indexing (LSI) aspects of the singular value decomposition (SVD), e.g., \cite{Bach, Ding, Dhillon0, Dhillon2, Drineas, Luxburg, Ng, Shi, Verma, Yu, Zha} for clustering aspect, and \cite{Alhabashneh, Berry2, Crain, Deerwester, Kolda, Kontostathis, Papadimitriou, Thorleuchter} for LSI aspect. In many literatures, eigenvector based clustering generally is known as spectral clustering; a technique involving computing eigenvectors of a graph's af\mbox{}finity matrix to find associations between related vertices. Because for any diagonalizable (the only case where eigendecomposition is defined) positive semi-definite matrix, eigenvectors and eigenvalues can be chosen so that are equivalent to singular vectors and singular values, a discussion on clustering aspect of the SVD can also cover spectral clustering\footnote{Diagonal shifting mechanism can be used to enforce positive semi-definiteness of a matrix \cite{Dhillon}.}. A review on some standard algorithms for spectral clustering can be found in a paper by Luxburg \cite{Luxburg}.

LSI is an indexing method that makes use of the truncated SVD to improve performance of an information retrieval (IR) system by also indexing terms that appear in related documents and weakening influence of terms that appear in unrelated documents \cite{Deerwester}. Accordingly, not only relevant documents that do not contain the query terms can be retrieved, but also irrelevant documents that contain query terms can be filtered out. These two simultaneous mechanisms generally can handle synonymy and polysemy problems to some extent.  

In this paper, clustering and LSI aspects of the SVD will be discussed, and a connnection between them will be drawn. In clustering part, first we will present theoretical supports on the using of the SVD in graph clustering including unipartite, bipartite, and directed graph clustering. Numerical results using synthetic and real datasets will be presented subsequently. In LSI part, first we will show how the SVD solves synonymy and polysemy problems in synthetic datasets. Then by recognizing that the SVD handles these problems by strengthening the grouping of related vertices, we will devise an LSI algorithm that can mimick this capability. Convergence analysis of the algorithm will be presented, and its retrieval performances will be compared to the SVD's using some standard datasets in LSI research subsequently.

A note on notation and term. Semantic graph refers to bipartite graph constructed from rectangular word-by-document matrix where the rows are the indexed words, the columns are the documents, and the entries are word frequencies in corresponding documents. $\mathbb{C}^{N\times K}$ denotes $N\times K$ complex matrix, $\mathbb{R}^{N\times K}$ denotes $N\times K$ real matrix, $\mathbb{R}_{+}^{N\times K}$ denotes $N\times K$ nonnegative real matrix, $\mathbb{B}_{+}^{N\times K}$ denotes $N\times K$ binary matrix, $k\in[1,K]$ denotes $k=1,\ldots,K$, and whenever complex matrix is concerned, transpose operation refers to conjugate transpose.

\section{The SVD}\label{ch2:svd}

The SVD is a matrix decomposition technique that factorizes a rectangular real or complex matrix into its left singular vectors, right singular vectors, and singular values. Some applications of the SVD are for approximating a matrix \cite{Eckart}, computing pseudoinverse of a matrix \cite{Golub}, determining the rank, range, and null space of a matrix \cite{Golub2}, and clustering \cite{Dhillon0,Drineas} among others.

The SVD of matrix $\mathbf{A}\in\mathbb{C}^{M\times N}$ with rank$(\mathbf{A})=r$ is defined with:
\begin{equation}
\mathbf{A} = \mathbf{U}\mathbf{\Sigma}\mathbf{V}^T,
\label{ch2:eq1}
\end{equation}
where $\mathbf{U}\in\mathbb{C}^{M\times M} = [\mathbf{u}_1,\ldots,\mathbf{u}_M]$ denotes a unitary matrix that contains left singular vectors of $\mathbf{A}$, $\mathbf{V}\in\mathbb{C}^{N\times N} = [\mathbf{v}_1,\ldots,\mathbf{v}_N]$ denotes a unitary matrix that contains right singular vectors of $\mathbf{A}$, and $\mathbf{\Sigma}\in\mathbb{R}_{+}^{M\times N}$ denotes a matrix that contains singular values of $\mathbf{A}$ along its diagonal with the diagonal entries $\sigma_1\ge\ldots\sigma_r>\sigma_{r+1}=\ldots=\sigma_{\min(M,N)}=0$ and zeros otherwise.

Rank-$K$ approximation of $\mathbf{A}$ using the SVD is defined with:
\begin{equation}
\mathbf{A}\approx\mathbf{A}_K = \mathbf{U}_K\mathbf{\Sigma}_K \mathbf{V}_K^T,
\label{ch2:eq2}
\end{equation}
where $K < r$, $\mathbf{U}_K$ and $\mathbf{V}_K$ contain the first $K$ columns of $\mathbf{U}$ and $\mathbf{V}$ respectively, and $\mathbf{\Sigma}_K$ denotes a $K\times K$ principal submatrix of $\mathbf{\Sigma}$. Eq.~\ref{ch2:eq2} is also known as the truncated SVD of $\mathbf{A}$, and according to the Eckart-Young theorem, $\mathbf{A}_K$ is the closest rank-$K$ approximation of $\mathbf{A}$ \cite{Eckart,Golub}.
\begin{theorem}[Eckart and Young \cite{Eckart,Golub}]\label{ch2:th1}
Let $\mathbf{A}_s$ ($s<r$) denotes rank-$s$ approximation of $\mathbf{A}$ using the SVD, then for all $\mathbf{B}\in\mathbb{C}^{M\times N}$ with rank$(\mathbf{B})=s$:
\begin{equation*}
\|\mathbf{A}-\mathbf{A}_s\|_F^2 \le \|\mathbf{A}-\mathbf{B}\|_F^2.
\end{equation*}
where $\|\mathbf{X}\|_F$ denotes Frobenius norm of $\mathbf{X}$.
\end{theorem}

Sometimes the SVD is being compared to eigendecomposition or also known as spectral decomposition, a technique that factorizes a square diagonalizable matrix into its eigenvectors and eigenvalues. Let $\mathbf{A}\in\mathbb{C}^{N\times N}$ denotes a diagonalizable matrix that has $N$ linearly independent eigenvectors, the eigendecomposition of $\mathbf{A}$ can be written as:
\begin{equation*}
\mathbf{A} = \mathbf{Q}\mathbf{\Lambda}\mathbf{Q}^{-1},
\end{equation*}
where $\mathbf{Q}=[\mathbf{q}_1,\ldots,\mathbf{q}_N]$, $\mathbf{q}_i$ denotes $i$-th eigenvector of $\mathbf{A}$, and $\mathbf{\Lambda}=\mathrm{diag}[$$\lambda_{i}$, $\ldots$, $\lambda_{N}]$ denotes a diagonal matrix with entry $\lambda_{i}$ is an eigenvalue corresponds to $\mathbf{q}_i$. If $\mathbf{A}$ is a Hermitian matrix, then the eigendecomposition can be written as:
\begin{equation}
\mathbf{A} = \mathbf{Q}\mathbf{\Lambda}\mathbf{Q}^T,
\label{ch2:eq4}
\end{equation}
where $\mathbf{Q}$ is a unitary matrix, i.e., $\mathbf{QQ}^T=\mathbf{Q}^T\mathbf{Q}=\mathbf{I}$, and $\mathbf{\Lambda}$ contains only real entries.

\section{Clustering Aspect of the SVD}\label{ch2:clusteringsvd}

Theoretical supports for clustering aspect of the SVD will be discussed in this section. We will first state the Ky Fan theorem \cite{Nakic,Zha}, a theorem that relates eigenvectors of a Hermitian matrix to trace maximization problem of the matrix. By showing that clustering problems of unipartite, bipartite, and directed graphs can be transformed into problems of finding eigenvectors of corresponding symmetric af\mbox{}finity matrices, we suggest that the Ky Fan theorem is a theoretical basis for spectral clustering. 

If the affinity matrix is also positive semi-definite, then eigendecomposition can be chosen to be equivalent to the SVD. Note that since diagonal shifting mechanism \cite{Dhillon} can be used to enforce positive semi-definiteness of an af\mbox{}finity matrix without altering any information in the corresponding graph, the Ky Fan theorem can also be considered as a theoretical support for clustering aspect of the SVD in graphs with symmetric affinity matrices. Further, by extending the theorem to rectangular matrix, we build a theoretical support for the direct use of the SVD in bipartite and directed graph clustering.

\subsection{The Ky Fan Theorem}\label{ch2:kyfan}
The following theorem states the Ky Fan theorem for Hermitian matrix. Without loss of generality, to simplify the presentation we will assume the matrix to be of full rank.
\begin{theorem}[Ky Fan \cite{Nakic,Zha}] \label{ch2:th2}
The optimal value of the following problem
\begin{equation}
\max_{\mathbf{X}^T\mathbf{X}=\mathbf{I}_K} \mathrm{tr}(\mathbf{X}^T\mathbf{H}\mathbf{X})
\label{ch2:eq5}
\end{equation}
is $\sum_{k=1}^{K}\lambda_k$ which is given by
\begin{equation*}
\mathbf{X} = [\mathbf{u}_1,\ldots,\mathbf{u}_K]\mathbf{Q},
\end{equation*}
where $\mathbf{H}\in\mathbb{C}^{N\times N}$ denotes a full rank Hermitian matrix with eigenvalues $\lambda_1\ge\ldots\ge\lambda_N\in\mathbb{R}$, $K\in[1,N]$, $\mathbf{X}\in\mathbb{C}^{N\times K}$ denotes a column orthogonal matrix, $\mathbf{I}_K$ denotes a $K\times K$ identity matrix, $\mathbf{u}_k\in\mathbb{C}^{N}$ denotes $k$-th eigenvector corresponds to $\lambda_k$, and $\mathbf{Q}\in\mathbb{C}^{K\times K}$ denotes an arbitrary unitary matrix.
\end{theorem}

Solution to eq.~\ref{ch2:eq5} is not unique since $\mathbf{X}$ remains equally good for any arbitrary rotation and reflection due to the existence of $\mathbf{Q}$. However, since $[\mathbf{u}_1,\ldots,\mathbf{u}_K]$ is one of the optimal solution, setting $\mathbf{X}=[\mathbf{u}_1,\ldots,\mathbf{u}_K]$ leads to the optimal value.

If $\mathbf{H}\leftarrow\mathbf{W}\in\mathbb{R}_{+}^{N\times N}$ where $\mathbf{W}$ denotes a symmetric af\mbox{}finity matrix induced from a graph, and $\mathbf{X}$ is constrained to be nonnegative while preserving the orthogonality, then eq.~\ref{ch2:eq5} turns into $K$-way graph cuts objective function. Therefore, the Ky Fan theorem can be viewed as a relaxed version of the graph cuts problem. This relationship suggests that the Ky Fan theorem is the theoretical support for spectral clustering.

In the next section, we will show how to modify unipartite, bipartite, and directed graph clustering objectives into trace maximization of symmetric matrices. By relaxing the nonnegativity constraints, clustering problems eventually become the tasks of computing the first $K$ eigenvectors of the matrices.

\subsection{Graph Clustering}\label{ch2:graphclustering}

Let $\mathcal{G}\left(\mathbf{A}\right)\equiv\mathcal{G}\left(\mathcal{V},\mathcal{E},\mathbf{A}\right)$ be the graph representation of matrix $\mathbf{A}$ with $\mathcal{V}$ denotes set of vertices and $\mathcal{E}$ denotes set of edges. $K$-way graph clustering is the problem of finding the best cuts on $\mathcal{G}\left(\mathbf{A}\right)$ that maximize within cluster associations, or equivalently, minimize inter cluster cuts to produce $K$ clusters of $\mathcal{V}$. Here we state two assumptions to allow graph cuts be employed in clustering.
\begin{assumption} \label{ch2:assump1}
Let $e_{ij}$ be an edge connecting vertex $v_i$ to $v_j$, and $|e_{ij}|$ be the weight of $e_{ij}$. We assume that edge weights describe similarities between vertices linearly, i.e., if $|e_{ij}| = n|e_{ik}|$ then $v_i$ is $n$ times more similar to $v_j$ than to $v_k$, and thus $v_i$ is $n$ times more likely to be grouped together with $v_j$ than with $v_k$. And zero weight means no similarity.
\end{assumption}

\begin{assumption} \label{ch2:assump2}
Graph clustering refers to hard clustering, i.e., $\{\mathcal{V}_k\}_{k=1}^{K}\subset\mathcal{V}$, $\cup_{k=1}^{K}\mathcal{V}_k=\mathcal{V}$, and $\mathcal{V}_k\cap\mathcal{V}_l=\emptyset$ $\forall k\neq l$.
\end{assumption}
\begin{claim} \label{ch2:prop1}
Assumption \ref{ch2:assump1} and \ref{ch2:assump2} lead to the grouping of similar vertices in $\mathcal{G}\left(\mathbf{A}\right)$. 
\end{claim}
\begin{proof}
Consider $\mathcal{G}\left(\mathbf{A}\right)$ to be clustered into $K$ groups by initial random assignments. Since assumption \ref{ch2:assump1} guarantees $|e_{ij}|$ to be comparable, and assumption \ref{ch2:assump2} guarantees each vertex to be assigned only to a single cluster, cluster assignment for $v_i$, $z_{ik}$, can be determined by finding a cluster's center that is most similar to $v_i$:
\begin{equation}
z_{ik} = \nut_{k}\,\max\left(\sum_{j_k=v_{j_k}\in\mathcal{V}_{k}}\frac{|e_{ij_k}|}{|\mathcal{V}_{k}|}\;\Bigg|\;k\in[1,K]\right),
\label{kmeans}
\end{equation}
where $|\mathcal{V}_{k}|$ denotes size of cluster $k$. Eq.~\ref{kmeans} is k-means clustering objective applied to $\mathcal{G}\left(\mathbf{A}\right)$, and therefore leads to the grouping of similar vertices.
\end{proof}

Note that assumption \ref{ch2:assump1} is an ideal situation which generally doesn't hold. For example, in bipartite graph representation of word-by-document matrix $\mathbf{A}$, similarity between two documents usually cannot be directly determined by computing distance between raw document vectors. Instead some preprocessing steps, e.g., feature selection and term weighting, are usually utilized to scale entries of the vectors. The preprocessing steps seem to be very crucial for obtaining good results \cite{Bach}, and many works, e.g.,  \cite{Bach, Jin, Wu, Jain}, were devoted to find more accurate similarity measure schemes. And even in the cases where similarities have been reflected by the weights in (almost) linear fashion, a normalization scheme on $\mathbf{A}$ generally is preferable to produce clusters with balanced sizes. In fact, normalized association/cuts are proven to of\mbox{}fer better results compared to their unnormalized counterparts, ratio association/cuts \cite{Shi,Yu,Dhillon}. 

Table \ref{ch2:table1} shows the most popular graph clustering objectives with the first two objectives are from the work of Dhillon et al.~\cite{Dhillon}. \emph{GWAssoc}/\emph{GWCuts} refer to general weighted association/cuts, \emph{NAssoc}/\emph{NCuts} refer to normalized association/cuts, and \emph{RAssoc}/\emph{RCuts} refer to ratio association/cuts. The definition of matrix $\mathbf{W}$, $\mathbf{\Phi}$, $\mathbf{L}$, and $\mathbf{D}$  will be given in the subsequent sections. Since all other objectives can be derived from \emph{GWAssoc} \cite{Dhillon}, we will only consider \emph{GWAssoc} for the rest of this paper.

\begin{table}[t]
  \begin{center}
    \caption{\small Graph clustering objectives.}
    \centering
    \footnotesize{
    \begin{tabular}{lll}
    \hline
   Objective & Af\mbox{}finity matrix & Weight matrix \\
    \hline
    \textit{GWAssoc} & $\mathbf{W}$ & $\mathbf{\Phi}$ \\
    \textit{GWCuts}  & $\mathbf{\Phi}-\mathbf{L}$ & $\mathbf{\Phi}$ \\
    \textit{NAssoc}  & $\mathbf{W}$ & $\mathbf{D}$ \\
    \textit{NCuts}   & $\mathbf{D}-\mathbf{L}$ & $\mathbf{D}$ \\
    \textit{RAssoc}  & $\mathbf{W}$ & $\mathbf{I}$ \\
    \textit{RCuts}   & $\mathbf{I}-\mathbf{L}$ & $\mathbf{I}$ \\
    \hline
    \end{tabular}}
    \label{ch2:table1}
  \end{center}
\end{table}

\subsubsection{Unipartite Graph Clustering} \label{ch2:unipartite}
We will first discuss unipartite graph because it is the framework for deriving a unified treatment for the three graphs. The following proposition summarizes a general unipartite graph clustering objective proposed by Dhillon et al.~\cite{Dhillon}.

\begin{proposition} \label{ch2:prop2}
Unipartite graph clustering can be stated in a trace maximization problem of a symmetric matrix.
\end{proposition}
\begin{proof}
Let $\mathbf{W}\in\mathbb{R}_{+}^{N\times N}$ be a symmetric af\mbox{}finity matrix induced from a unipartite graph, $K$-way partitioning on $\mathcal{G}\left(\mathbf{W}\right)$ using \textit{GWAssoc} can be found by:
\begin{equation*}
\max\;J_u = \frac{1}{K}\sum_{k=1}^{K}\frac{z^T_k\mathbf{W}z_k}{z^T_k\mathbf{\Phi}z_k}
\end{equation*}
where $\mathbf{\Phi}\in\mathbb{R}^{N\times N}_{+}$ denotes a diagonal matrix with entry $\Phi_{ii}$ denotes weight of $v_i$, and $\mathbf{z}_k\in\mathbb{B}^{N}_{+}$ denotes a binary indicator vector for cluster $k$ with its $i$-th entry is 1 if $v_i$ in cluster $k$, and 0 otherwise. The above objective can be rewritten in a trace maximization as:
\begin{align}
\max\;J_u &= \frac{1}{K}\text{tr}\left(\frac{\mathbf{Z}^T\mathbf{W}\mathbf{Z}}{\mathbf{Z}^T\mathbf{\Phi}\mathbf{Z}}\right) \nonumber \\
        &= \frac{1}{K}\text{tr}\left(\bar{\mathbf{Z}}^T\mathbf{\Phi}^{-1/2}\mathbf{W}\mathbf{\Phi}^{-1/2}\bar{\mathbf{Z}}\right)
\label{ch2:eq6}
\end{align}
where $\mathbf{Z}\in\mathbb{B}^{N\times K}_{+}=[\mathbf{z}_1,\ldots,\mathbf{z}_K]$ denotes clustering indicator matrix, and $\bar{\mathbf{Z}}\in\mathbb{R}^{N\times K}_{+}=\mathbf{Z}/\left(\mathbf{Z}^T\mathbf{Z}\right)^{1/2}$ denotes its orthonormal version.
\end{proof}

By relaxing the nonnegativity constraints, i.e., allowing $\bar{\mathbf{Z}}$ to contain negative values while preserving the orthonormality, by the Ky Fan theorem, a global optimum of $J_u$ can be obtained by:
\begin{equation}
\mathbf{\hat{Z}} = [\mathbf{u}_1,\ldots,\mathbf{u}_K],
\label{ch2:eq7}
\end{equation}
where $\mathbf{u}_1,\ldots,\mathbf{u}_K\in\mathbb{C}^N$ denote the first $K$ eigenvectors of $\mathbf{\Phi}^{-1/2}\mathbf{W}\mathbf{\Phi}^{-1/2}$ and $\mathbf{\hat{Z}}\in\mathbb{C}^{N\times K}$ denotes a relaxed version of $\mathbf{\bar{Z}}$. 

As shown, eq.~\ref{ch2:eq7} presents a tractable solution for NP-hard problem in eq.~\ref{ch2:eq6}. The reverse operation, i.e., inferring an approximate to $\mathbf{Z}$ from $\mathbf{\hat{Z}}$ can be conducted by applying k-means clustering on rows of $\mathbf{\hat{Z}}$ as will be shown later in the last step of algorithm \ref{ch2:alg1}.

The \emph{GWAsssoc} objective in eq.~\ref{ch2:eq6} can be replaced by any objective in table \ref{ch2:table1} by substituting $\mathbf{W}$ and $\mathbf{\Phi}$ with corresponding af\mbox{}finity and weight matrices. Note that $\mathbf{I}$ denotes the identity matrix, $\mathbf{D}\in\mathbb{R}^{N\times N}_{+}$ denotes a diagonal matrix with entries defined as $D_{ii}=\sum_j W_{ij}$, and $\mathbf{L}=\mathbf{D}-\mathbf{W}$ denotes the Laplacian of $\mathcal{G}\left(\mathbf{W}\right)$.

\subsubsection{Bipartite Graph Clustering} \label{ch2:bipartite}

A bipartite graph is a graph with two types of vertices each belongs to a different independent set and every edge can only connect vertex pairs from different sets. In this paper, we refer a bipartite dataset as a collection of items that are characterized by some shared features. A feature-by-item rectangular af\mbox{}finity matrix $\mathbf{A}\in\mathbb{R}^{M\times N}_{+}$ can then be induced to represent the dataset. 

Bipartite graph clustering can be applied in direct and indirect ways. The former applies graph clustering directly to $\mathcal{G}\left(\mathbf{A}\right)$ resulting in partitions that contain both item and feature vertices. The latter first transforms $\mathcal{G}\left(\mathbf{A}\right)$ into an equivalent unipartite graph (either item or feature graph) by calculating similarities between vertex pairs from either item or feature set, and then applies graph clustering to this transformed graph. As will be shown, both lead to symmetric af\mbox{}finity matrices, and accordingly \emph{GWAssoc} can be applied equivalently as in the unipartite graph clustering.

\paragraph{Direct Treatment.} If \emph{GWAssoc} is applied directly to a bipartite graph, similar items will be grouped together with relevant features. This is known as simultaneous feature and item clustering or \emph{co-clustering}. The following proposition describes co-clustering.

\begin{proposition} \label{ch2:prop3}
Bipartite graph co-clustering can be stated in a trace maximization problem of a symmetric matrix.
\end{proposition}
\begin{proof}
Let $\mathbf{A}\in\mathbb{R}^{M\times N}_{+}$ be a feature-by-item matrix representing a bipartite graph, then a symmetric matrix induced from the graph can be written as:
\begin{equation*}
\mathbf{M} = \left[
 \begin{array}{cc}
  \mathbf{0}   & \mathbf{A} \\
  \mathbf{A}^T & \mathbf{0} \\
 \end{array} \right]\in\mathbb{R}^{P\times P}_{+}.
\end{equation*}
Taking \textit{GWAssoc} as the objective, $K$-way co-clustering can be found by:
\begin{align}
\max\;J_b &= \frac{1}{K}\sum_{k=1}^{K}\frac{z^T_k\mathbf{M}z_k}{z^T_k\mathbf{\Phi}z_k} \nonumber \\
          &= \frac{1}{K}\text{tr}\left(\bar{\mathbf{Z}}^T\mathbf{\Phi}^{-1/2}\mathbf{M}\mathbf{\Phi}^{-1/2}\bar{\mathbf{Z}}\right),
\label{ch2:eq9}
\end{align}
where $\mathbf{\Phi}\in\mathbb{R}^{P\times P}_{+}$, $\mathbf{z}_k\in\mathbb{B}^{P}_{+}$, and $\bar{\mathbf{Z}}\in\mathbb{R}^{P\times K}_{+}$ are defined equivalently as in proposition \ref{ch2:prop2}.
\end{proof}

By relaxing the nonnegativity constraints in $\bar{\mathbf{Z}}$, the optimal value of eq.~\ref{ch2:eq9} can be found by computing the first $K$ eigenvectors of $\mathbf{\Phi}^{-1/2}\mathbf{M}\mathbf{\Phi}^{-1/2}$.

Instead of constructing $\mathbf{M}$ which is bigger and sparser than $\mathbf{A}$, the following theorem provides a way to co-cluster bipartite graph directly from $\mathbf{A}$.

\begin{theorem} \label{ch2:th3}
A relaxed solution to the bipartite graph co-clustering problem in eq.~\ref{ch2:eq9} can be found by computing left and right singular vectors of a normalized version of $\mathbf{A}$.
\end{theorem}

\begin{proof}
Let
\begin{equation*}
\mathbf{\bar{Z}}=\left[\begin{array}{c}\mathbf{\bar{X}}\\\mathbf{\bar{Y}}\end{array}\right]\,\mathrm{,}\;\mathrm{and}\;
\mathbf{\Phi}=\left[\begin{array}{cc}\mathbf{\Phi}_{1} & \mathbf{0}\\
\mathbf{0} & \mathbf{\Phi}_{2}\end{array}\right]
\end{equation*}
be rearranged into two smaller matrices. Then, eq.~\ref{ch2:eq9} can be rewritten as:
\begin{equation}
\max\;J_b = \frac{1}{K}\text{tr}\left(\left[\begin{array}{c}\mathbf{\bar{X}}\\\mathbf{\bar{Y}}\end{array}\right]^T\underbrace{\left[\begin{array}{cc}\mathbf{0} & \mathbf{\bar{A}}\\
\mathbf{\bar{A}}^T & \mathbf{0}\end{array}\right]}_{\mathbf{\bar{M}}}\left[\begin{array}{c}\mathbf{\bar{X}}\\\mathbf{\bar{Y}}\end{array}\right]\right),
\label{ch2:eq10}
\end{equation}
where $\mathbf{\bar{A}}=\mathbf{\Phi}_{1}^{-1/2}\mathbf{A}\mathbf{\Phi}_{2}^{-1/2}$ denotes the normalized version of $\mathbf{A}$. Denoting $\mathbf{\hat{X}}\in\mathbb{C}^{M\times K}$ and $\mathbf{\hat{Y}}\in\mathbb{C}^{N\times K}$ as a relaxed version of $\mathbf{\bar{X}}$ and $\mathbf{\bar{Y}}$ respectively, by the Ky Fan theorem, a global optimum solution to eq.~\ref{ch2:eq10} is given by the first $K$ eigenvectors of $\mathbf{\bar{M}}$:
\begin{equation*}
\left[\begin{array}{c}\mathbf{\hat{X}}\\\mathbf{\hat{Y}}\end{array}\right]=
\left[\begin{array}{c}\mathbf{\hat{x}}_1,\ldots,\mathbf{\hat{x}}_K\\
\mathbf{\hat{y}}_1,\ldots,\mathbf{\hat{y}}_K\end{array}\right].
\end{equation*}
Therefore, 
\begin{equation*}
\left[\begin{array}{cc}\mathbf{0} & \mathbf{\bar{A}}\\
\mathbf{\bar{A}}^T & \mathbf{0}\end{array}\right]\left[\begin{array}{c}\mathbf{\hat{x}}_k\\
\mathbf{\hat{y}}_k\end{array}\right] =
\lambda_k\left[\begin{array}{c}\mathbf{\hat{x}}_k\\
\mathbf{\hat{y}}_k\end{array}\right],\;\forall k\in[1,K],
\end{equation*}
where $\lambda_k$ denotes $k$-th eigenvalue of $\mathbf{\bar{M}}$. Then,
\begin{align}
\mathbf{\bar{A}}\mathbf{\hat{y}}_k &= \lambda_k\mathbf{\hat{x}}_k,\,\; \text{and} \label{ch2:eq11}\\
\mathbf{\bar{A}}^T\mathbf{\hat{x}}_k &= \lambda_k\mathbf{\hat{y}}_k,
\label{ch2:eq12}
\end{align}
which are the definitions of singular values and singular vectors of $\mathbf{\bar{A}}$, where $\mathbf{\hat{y}}_k$ and $\mathbf{\hat{x}}_k$ denote left and right singular vectors associated with singular value $\lambda_k$. Thus, the relaxed solution to the problem in eq.~\ref{ch2:eq9} can be found by computing the first $K$ left and right singular vectors of $\mathbf{\bar{A}}$.
\end{proof}

\paragraph{Indirect Treatment.} There are cases where the data points are inseparable in the original space or clustering can be done more effectively by first transforming $\mathbf{A}$ into a corresponding symmetric matrix $\mathbf{\dot{A}}\in\mathbb{R}_{+}^{N\times N}$ (we assume item clustering here, feature clustering can be treated similarly), and then applying graph clustering on $\mathcal{G}(\mathbf{\dot{A}})$.

There are two common approaches to learn $\mathbf{\dot{A}}$ from $\mathbf{A}$. The first approach is to use kernel functions. Table \ref{ch2:table2} enlists some widely used kernel functions with $\mathbf{a}_i$ denotes $i$-th column of $\mathbf{A}$, and the unknown parameters ($c,d,\alpha$, and $\theta$) are either determined based on previous experiences or learned directly from sample datasets. In section \ref{ch2:synteticdata}, we will provide some examples on how to solve bipartite graph clustering using kernel approach.

The second approach is to make no assumption about data domain nor possible similarity structure between item pairs. $\mathbf{\dot{A}}$ is learned directly from the data, thus avoiding some inherent problems associated with the first approach, e.g., (1) no standard in choosing kernel functions and (2) similarities between item pairs are computed independently without considering interactions between items. Some recent works on the second approach can be found in \cite{Jin, Wu, Jain}. 

If asymmetric metrics like Bregman divergences are used as kernel functions, $\mathbf{\dot{A}}$ will be asymmetric. Accordingly, $\mathcal{G}(\mathbf{\dot{A}})$ will be a directed graph, and therefore it must be treated as a directed graph.

\begin{table}[t]
 \renewcommand{\arraystretch}{1.3}
  \begin{center}
    \caption{\small Examples of popular kernel functions \cite{Dhillon}.}
    \centering
    \footnotesize{
    \begin{tabular}{ll}
    \hline
    Polynomial kernel & $\kappa(\mathbf{a}_i,\mathbf{a}_j)=(\mathbf{a}_i\cdot\mathbf{a}_j+c)^d$ \\
    Gaussian kernel   & $\kappa(\mathbf{a}_i,\mathbf{a}_j)=\mathrm{exp}(-\|\mathbf{a}_i-\mathbf{a}_j\|^2/2\alpha^2)$ \\
    Sigmoid kernel    & $\kappa(\mathbf{a}_i,\mathbf{a}_j)=\mathrm{tanh}(c(\mathbf{a}_i\cdot\mathbf{a}_j)+\theta)$ \\
    \hline
    \end{tabular}}
    \label{ch2:table2}
  \end{center}
\end{table}

\subsubsection{Directed Graph Clustering} \label{ch2:directed}

Research on directed graph clustering mainly comes from the study on complex networks. Dif\mbox{}ferent from conventional method of ignoring edge directions, complex network researchers preserve this information in the proposed methods. As shown in \cite{Leicht, Kim}, accomodating this information can be very useful in improving clustering quality. And in some cases, ignoring edge directions can lead to the clusters detection failure \cite{Kim1}.

Directed graph clustering usually is done by mapping the original square af\mbox{}finity matrix into another square matrix which entries have been adjusted to emphasize the importance of edge directions. Some mapping functions can be found in, e.g., \cite{Leicht, Kim, Kim1}. To make use of the available clustering methods for unipartite graph, some works proposed constructing a symmetric matrix representing the directed graph without ignoring edge directions \cite{Leicht, Kim1}.

Here we will describe directed graph clustering by following previous discussions on unipartite and bipartite graph clustering.

\begin{proposition} \label{ch2:prop5}
Directed graph clustering can be stated in a trace maximization problem of a symmetric af\mbox{}finity matrix.
\end{proposition}
\begin{proof}
Let $\mathbf{B}\in\mathbb{R}_{+}^{N\times N}$ denotes an af\mbox{}finity matrix induced from the directed graph, and $\mathbf{\Phi}_i$ and $\mathbf{\Phi}_o$ denote diagonal weight matrices associated with indegree and outdegree of vertices in $\mathcal{G}\left(\mathbf{B}\right)$ respectively. Then, a diagonal weight matrix of $\mathcal{G}\left(\mathbf{B}\right)$ can be defined with:
\begin{equation*}
\mathbf{\Phi}_{io} = \sqrt{\mathbf{\Phi}_{i}\mathbf{\Phi}_{o}}.
\end{equation*}
Since both rows and columns of $\mathbf{B}$ correspond to the same set of vertices with the same order, the row and column clustering indicator matrices must be the same, matrix $\mathbf{\bar{Z}}$. By using \emph{GWAssoc}, $K$-way clustering on $\mathcal{G}\left(\mathbf{B}\right)$ and $\mathcal{G}\left(\mathbf{B}^T\right)$ can be computed by:
\begin{align*}
\max\;J_{d1} &= \frac{1}{K}\text{tr}\left(\bar{\mathbf{Z}}^T\mathbf{\Phi}_{io}^{-1/2}\mathbf{B}\mathbf{\Phi}_{io}^{-1/2}\bar{\mathbf{Z}}\right),\;\text{and} 
\\
\max\;J_{d2} &= \frac{1}{K}\text{tr}\left(\bar{\mathbf{Z}}^T\mathbf{\Phi}_{io}^{-1/2}\mathbf{B}^T\mathbf{\Phi}_{io}^{-1/2}\bar{\mathbf{Z}}\right)
\end{align*}
respectively. By adding the two objectives above, we obtain:
\begin{equation}
\max\;J_d = \frac{1}{K}\text{tr}\left(\bar{\mathbf{Z}}^T\mathbf{\Phi}_{io}^{-1/2}\left(\mathbf{B}+\mathbf{B}^T\right)\mathbf{\Phi}_{io}^{-1/2}\bar{\mathbf{Z}}\right),
\label{ch2:eq13}
\end{equation}
which is a trace maximization problem of a symmetric matrix.
\end{proof}

Directed graph clustering raises an interesting issue in weight matrix formulation which doesn't appear in unipartite and bipartite graph cases as the edges are undirected. As explained in \cite{Dhillon}, $\mathbf{\Phi}$ is introduced with two purposes: (1) to provide a general form of graph cuts objective which other objectives can be derived from it, and (2) to provide compatibility with weighted kernel k-means objective so that eigenvector-free k-means algorithm can be utilized to solve graph clustering problem.

However, as information on edge direction appears, defining a weight for each vertex is no longer adequate. To see the reason, let's apply \emph{NAssoc} to $\mathcal{G}\left(\mathbf{B}\right)$ and $\mathcal{G}\left(\mathbf{B}^T\right)$. By using table \ref{ch2:table1}:
\begin{align*}
\max\;J_{d1} &= \frac{1}{K}\text{tr}\left(\bar{\mathbf{Z}}^T\mathbf{D}^{-1/2}\mathbf{B}\mathbf{D}^{-1/2}\bar{\mathbf{Z}}\right)\;\,\text{and} 
\\
\max\;J_{d2} &= \frac{1}{K}\text{tr}\left(\bar{\mathbf{Z}}^T\mathbf{D^{\ast}}^{-1/2}\mathbf{B}^T\mathbf{D^{\ast}}^{-1/2}\bar{\mathbf{Z}}\right),
\end{align*}
where $D$ and $D^{\ast}$ denote diagonal weight matrices with $D_{ii}=\sum_jB_{ij}$ and $D_{ii}^{\ast}=\sum_iB_{ij}$ respectively. Hence, $J_{d1}+J_{d2}$ won't end up in a trace maximization of a symmetric matrix as in eq.~\ref{ch2:eq13}. 

This motivates us to define a more general form of weight matrix, $\mathbf{\Phi}_{io}$, which allows directed graph clustering be stated in a trace maximization of a symmetric matrix, yet still turns into $\mathbf{\Phi}$ if corresponding af\mbox{}finity matrix is symmetric. 

In case of \emph{NAssoc} and \emph{NCuts}, $\mathbf{\Phi}_{i}$ and $\mathbf{\Phi}_{o}$ are defined as:
\begin{align*}
\mathbf{\Phi}_i&=\text{diag}\left(\sum_{i}B_{i1},\ldots,\sum_{i}B_{iN}\right)\;\,\text{and} \\
\mathbf{\Phi}_o&=\text{diag}\left(\sum_{j}B_{1j},\ldots,\sum_{j}B_{Nj}\right).
\end{align*}
And for \emph{RAssoc} and \emph{RCuts}: $\mathbf{\Phi}_i = \mathbf{\Phi}_o = \mathbf{I}$.

\subsection{Extension to the Ky Fan Theorem} \label{extension}
Theorem \ref{ch2:th3} implies an extension to the Ky Fan theorem for more general rectangular complex matrix. The following theorem states a generalized version of the Ky Fan theorem. And without loss of generality, we will assume the matrix to be of full rank to simplify the presentation.
\begin{theorem} \label{ch2:th6}
The optimal value of the following problem:
\begin{equation}
\max_{\mathbf{X}^T\mathbf{X}=\mathbf{Y}^T\mathbf{Y}=\mathbf{I}_K} \mathrm{tr}(\mathbf{X}^T\mathbf{R}\mathbf{Y}),
\label{ch2:eq14}
\end{equation}
is $\sum_{k=1}^{K}\sigma_k$ which is given by
\begin{align*}
\mathbf{X} &= [\mathbf{x}_1,\ldots,\mathbf{x}_K]\mathbf{Q},\;\,\text{and}
\\
\mathbf{Y} &= [\mathbf{y}_1,\ldots,\mathbf{y}_K]\mathbf{Q}
\end{align*}
where $\mathbf{R}\in\mathbb{C}^{M\times N}$ denotes a full rank rectangular complex matrix with singular values $\sigma_1\ge\ldots\ge\sigma_{\min(M,N)}>0$, $K\in[1,\min(M,N)]$, $\mathbf{X}\in\mathbb{C}^{M\times K}$ and $\mathbf{Y}\in\mathbb{C}^{N\times K}$ denote column orthogonal matrices, $\mathbf{x}_k$ and $\mathbf{y}_k$ ($k\in[1,K]$) respectively denote $k$-th left and right singular vectors  correspond to $\sigma_k$, and $\mathbf{Q}\in\mathbb{C}^{K\times K}$ denotes an arbitrary unitary matrix.
\end{theorem}

\begin{proof}
Eq.~\ref{ch2:eq14} can be rewritten as:
\begin{equation*}
\max_{\mathbf{X}^T\mathbf{X}=\mathbf{Y}^T\mathbf{Y}=\mathbf{I}_K}\frac{1}{2} \mathrm{tr}\left(\left[\begin{array}{c}\mathbf{X}\\\mathbf{Y}\end{array}\right]^T\underbrace{\left[\begin{array}{cc}\mathbf{0} & \mathbf{R}\\
\mathbf{R}^T & \mathbf{0}\end{array}\right]}_{\mathbf{\Psi}}\left[\begin{array}{c}\mathbf{X}\\\mathbf{Y}\end{array}\right]\right).
\end{equation*}
Since $\mathbf{\Psi}$ is a Hermitian matrix, by the Ky Fan theorem a global optimum solution is given by the first $K$ eigenvectors of $\mathbf{\Psi}$:
\begin{equation*}
\left[\begin{array}{c}\mathbf{X}\\\mathbf{Y}\end{array}\right]=
\left[\begin{array}{c}\mathbf{x}_1,\ldots,\mathbf{x}_K\\
\mathbf{y}_1,\ldots,\mathbf{y}_K\end{array}\right].
\end{equation*}
By following proof of theorem \ref{ch2:th3}, it can be shown that $\mathbf{x}_1,\ldots,\mathbf{x}_K$ and $\mathbf{y}_1,$ $\ldots,$ $\mathbf{y}_K$ are the first $K$ left and right singular vectors of $\mathbf{R}$.
\end{proof}

\subsection{Experimental Results} \label{ch2:results1}

The experiments were conducted using a notebook with 1.86 GHz Intel processor and 2 GB RAM. All algorithms were developed in GNU Octave under linux platform. We used synthetic datasets to describe the use of the SVD in clustering graphs with af\mbox{}finity matrices have been transformed into corresponding symmetrix matrices, and real datasets to illustrate the direct use of the SVD in bipartite graph clustering.

\subsubsection{Synthetic Datasets}\label{ch2:synteticdata}
\begin{figure}
 \begin{center}
  \subfigure[$\alpha=0.05$]{
   \includegraphics[width=0.45\textwidth]{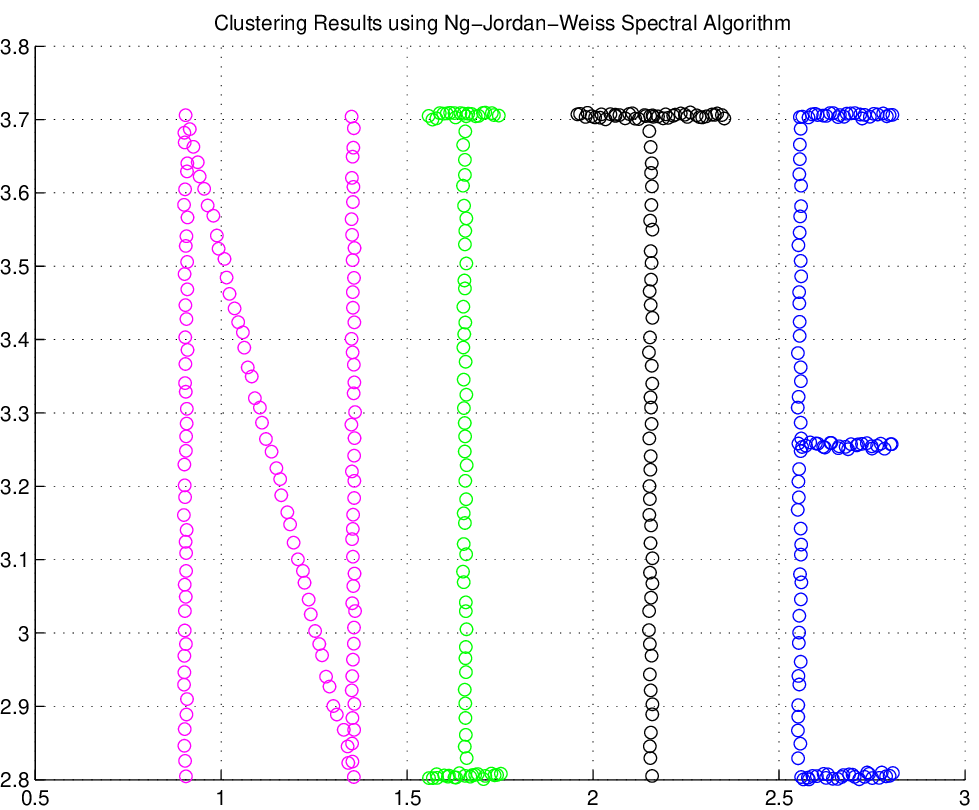}
   \label{ch2:fig1a}
  }
  \subfigure[$\alpha=0.1$]{
   \includegraphics[width=0.45\textwidth]{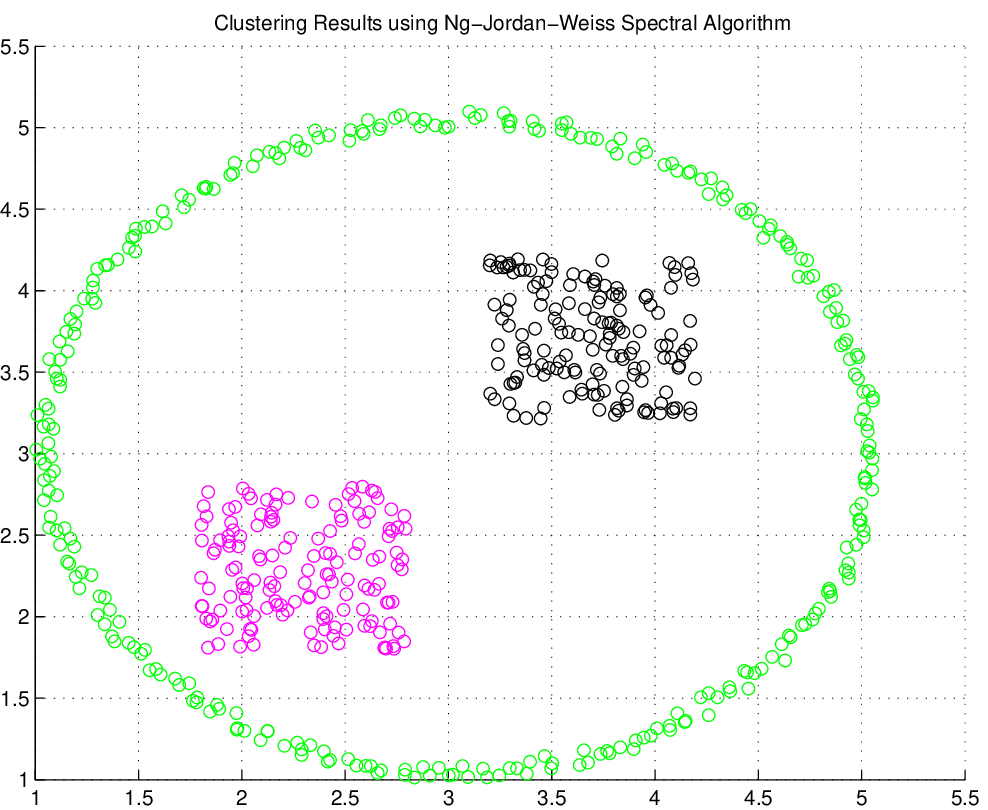}
   \label{ch2:fig1b}
  }
\\
  \subfigure[$\alpha=0.1$]{
   \includegraphics[width=0.45\textwidth]{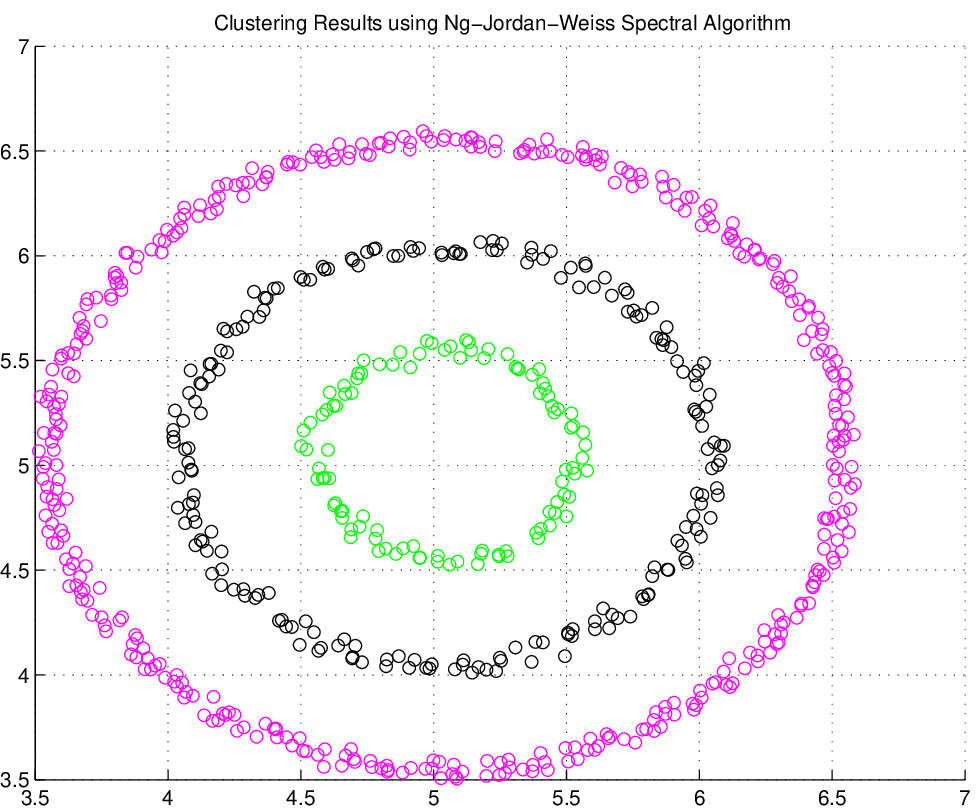}
   \label{ch2:fig1c}
  }
  \subfigure[$\alpha=0.2$]{
   \includegraphics[width=0.45\textwidth]{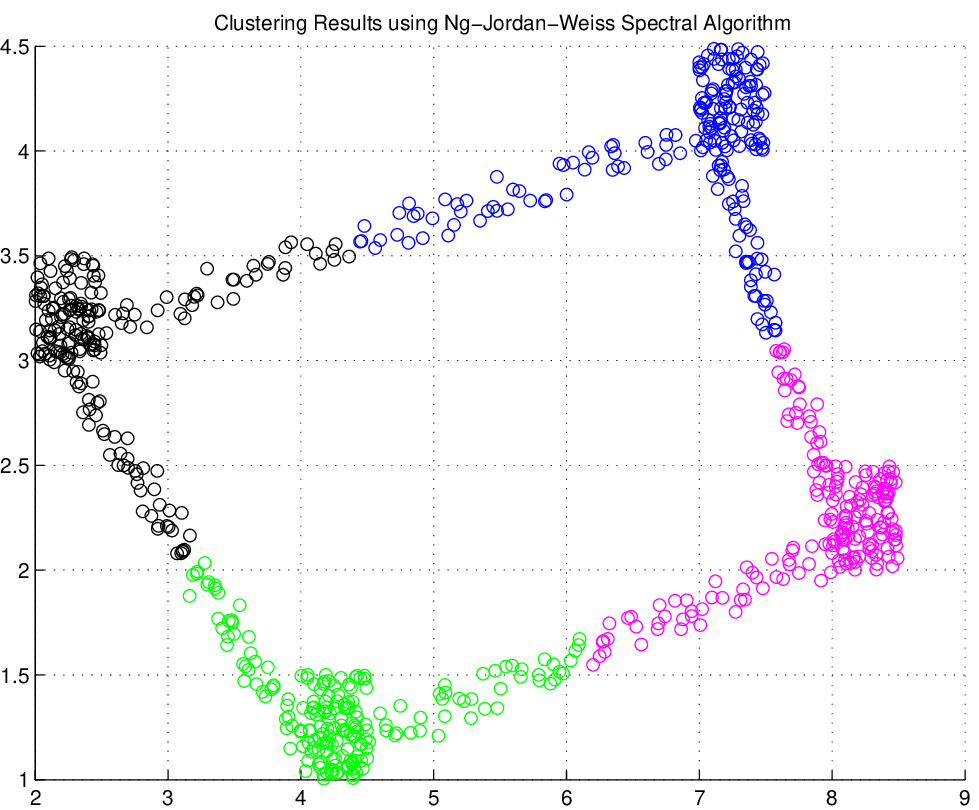}
   \label{ch2:fig1d}
  }
  \caption{Clustering synthetic datasets using algorithm \ref{ch2:alg1}. As shown, by choosing an appropriate $\alpha$ value for each dataset, the algorithm successfully found the correct clusters.}
  \label{ch2:fig1}
 \end{center}
\end{figure}

As shown in figure \ref{ch2:fig1}, the synthetic datasets are linearly inseparable with rectangular af\mbox{}finity matrices $\mathbf{A}$s. A common practice to cluster this kind of datasets is to use kernel approach to make the data points linearly separable in the tranformed space. Here, Gaussian kernel was chosen to transform $\mathbf{A}$s into corresponding symmetric $\mathbf{\dot{A}}$s that are expected to be better conditioned for clustering. According to the Ky Fan theorem, $K$-way cuts on the graph can be found by calculating $K$ eigenvectors correspond to the $K$ largest eigenvalues of $\mathbf{\dot{A}}$. Because the eigenvectors were mixed-signed, clustering indicators were induced from the eigenvectors using k-means. Algorithm \ref{ch2:alg1} describes the clustering procedure, and figure \ref{ch2:fig1} shows the results. Note that algorithm \ref{ch2:alg1} is spectral clustering algorithm proposed by Ng et al.~\cite{Ng}.

\begin{algorithm}[H]
\caption{Spectral clustering algorithm proposed by Ng et al.~\cite{Ng}.}
\label{ch2:alg1}
\begin{algorithmic}
\STATE \begin{enumerate}
\item Input: Rectangular data matrix $\mathbf{A}\in\mathbb{R}^{M\times N}_{+}$ with $N$ data points, \#cluster $K$, and Gaussian kernel parameter $\alpha$.
\item Construct symmetric af\mbox{}finity matrix $\mathbf{\dot{A}}\in\mathbb{R}^{N\times N}_{+}$ from $\mathbf{A}$ by using Gaussian kernel.
\item Normalize $\mathbf{\dot{A}}$ by $\mathbf{\dot{A}}\leftarrow\mathbf{D}^{-1/2}\mathbf{\dot{A}}\mathbf{D}^{-1/2}$ where $\mathbf{D}$ is a diagonal matrix with $D_{ii}=\sum_j\dot{A}_{ij}$.
\item Compute $K$ eigenvectors that correspond to the $K$ largest eigenvalues of $\mathbf{\dot{A}}$, and form $\mathbf{\hat{X}}\in\mathbb{R}^{N\times K}=[\mathbf{\hat{x}}_1,\ldots,\mathbf{\hat{x}}_K]$, where $\mathbf{\hat{x}}_k$ is the $k$-th eigenvector.
\item Normalize every row of $\mathbf{\hat{X}}$, i.e., $X_{ij}\leftarrow X_{ij}/(\sum_j X_{ij}^2)^{1/2}$.
\item Apply k-means clustering on rows of $\mathbf{\hat{X}}$ to obtain clustering indicator matrix $\mathbf{X}\in\mathbb{B}^{N\times K}_{+}$.
\end{enumerate}
\end{algorithmic}
\end{algorithm}

\subsubsection{Real Datasets}\label{ch2:realdata}
To illustrate the direct use of the SVD in bipartite graph clustering, the Reuters-21578 data corpus\footnote{http://kdd.ics.uci.edu/databases/reuters21578/reuters21578.html} was used. The detailed discussion on the dataset and preprocessing steps can be found in \cite{Mirzal2}.

We formed test datasets by combining top 2, 4, 6, 8, 10, and 12 classes from the corpus. Table \ref{ch2:table3} summarizes the statistics of these test datasets, where \#doc, \#word, \%nnz, max, and min denote the number of documents, the number of words, percentage of nonzero entries, maximum cluster size, and minimum cluster size respectively. Table \ref{ch2:table4} gives sizes (\#doc) of these top 12 classes. And algorithm \ref{ch2:alg2} outlines the clustering procedure.

\begin{table}
  \begin{center}
    \caption{Statistics of the test datasets}
    \centering
    \begin{tabular}{lrrrrr}
    \hline
    The data &\#doc & \#word & \%nnz & max & min \\
    \hline
    Reuters2    & 6090 & 8547  & 0.363 & 3874 & 2216 \\
    Reuters4    & 6797 & 9900  & 0.353 & 3874 & 333 \\
    Reuters6    & 7354 & 10319 & 0.347 & 3874 & 269 \\
    Reuters8    & 7644 & 10596 & 0.340 & 3874 & 144 \\
    Reuters10   & 7887 & 10930 & 0.336 & 3874 & 114 \\
    Reuters12   & 8052 & 11172 & 0.333 & 3874 &  75 \\
    \hline
    \end{tabular}
    \label{ch2:table3}
  \end{center}
\end{table}

\begin{table}
  \begin{center}
    \caption{Sizes of the top 12 topics}
    \centering
    \begin{tabular}{rrrrrrrrrrrr}
    \hline
    class &    1 &    2 &   3 &   4 &   5 &   6 \\ 
    \#doc & 3874 & 2216 & 374 & 333 & 288 & 269 \\ \hline
    class &   7 &   8 &   9 &  10 & 11 & 12 \\ 
    \#doc & 146 & 144 & 129 & 114 & 90 & 75 \\ \hline
    \end{tabular}
    \label{ch2:table4}
  \end{center}
\end{table}

\begin{algorithm}[H]
\caption{Bipartite graph clustering using the SVD.}
\label{ch2:alg2}
\begin{algorithmic}
\STATE \begin{enumerate}
\item Input: Rectangular word-by-document matrix $\mathbf{A}\in\mathbb{R}^{M\times N}_{+}$, and \#cluster $K$.
\item Normalize $\mathbf{A}$ by: $\mathbf{A} \leftarrow \mathbf{AD}^{-1/2}$ where $\mathbf{D}=\text{diag}( \mathbf{A}^T\mathbf{A}\mathbf{e})$.
\item Compute the first $K$ right singular vectors of $\mathbf{A}$, and form $\mathbf{\hat{V}}\in\mathbb{R}^{N\times K}=[\mathbf{\hat{v}}_1,\ldots,\mathbf{\hat{v}}_K]$, where $\mathbf{\hat{v}}_k$ is the $k$-th right singular vector.
\item Normalize every row of $\mathbf{\hat{V}}$, i.e., $V_{ij}\leftarrow V_{ij}/(\sum_j V_{ij}^2)^{1/2}$.
\item Apply k-means clustering on rows of $\mathbf{\hat{V}}$ to obtain clustering indicator matrix $\mathbf{V}\in\mathbb{B}^{N\times K}_{+}$.
\end{enumerate}
\end{algorithmic}
\end{algorithm}

To assess clustering quality, four standard metrics were used: mutual information (MI), entropy, purity, and Fmeasure. Definitions of these metrics have been outlined in \cite{Mirzal2}. In summary, for MI, purity, and Fmeasure, the higher the values the better the clustering quality. And for entropy, the smaller the values the better the clustering quality.

To evaluate performances of the SVD, Lee and Seung's nonnegative matrix factorization (NMF) algorithm \cite{Lee2} and k-means were used for comparison purpose. However, due to poor performance of k-means, only results of the NMF are shown. And because of the nonconvexity of the NMF, results for the NMF were averaged over 10 trials. Algorithm \ref{ch3:alg3} describes the standard clustering mechanism by using NMF as used in ref.~\cite{Xu, Shahnaz, JKim, TLi, Ding1, Ding2, Yoo2, Gu}\footnote{There is no need to normalize columns of $\mathbf{C}$ because in the NMF clustering assignment of $n$-th document is determined by the largest entry in corresponding column of $\mathbf{C}$.}. Table \ref{ch2:table5} shows the results. As shown, in general the SVD outperformed the NMF except for Reuters12 dataset. These results were expected because clustering indicator matrix in the NMF was determined directly by the most positive entry in $K$-subspace for each data point so that the NMF is limited for clustering linearly separable data points.

\begin{algorithm}[H]
\caption{Document clustering using NMF.}
\label{ch3:alg3}
\begin{algorithmic}
\STATE \begin{enumerate}
\item Input: Rectangular word-by-document matrix $\mathbf{A}\in\mathbb{R}^{M\times N}_{+}$, and \#cluster $K$.
\item Normalize $\mathbf{A}$ by: $\mathbf{A} \leftarrow \mathbf{AD}^{-1/2}$ where $\mathbf{D}=\text{diag}( \mathbf{A}^T\mathbf{A}\mathbf{e})$.
\item Compute coefficient matrix $\mathbf{C}\in\mathbb{R}^{K\times N}_{+}$ using Lee and Seung's algorithm \cite{Lee2} so that $\mathbf{A}\approx\mathbf{BC}$ where $\mathbf{B}\in\mathbb{R}^{M\times K}_{+}$ denotes basis matrix.
\item Compute clustering assignment of $n$-th document by: $x_n\longleftarrow\nut_k\max\mathbf{c}_n,\;\forall n$.
\end{enumerate}
\end{algorithmic}
\end{algorithm}

\begin{table}
 \begin{center}
   \caption{Clustering performance comparison.}
   \centering
   \begin{tabular}{lrrrrrrrr}
   \hline
     & \multicolumn{2}{c}{MI} & \multicolumn{2}{c}{Entropy} & \multicolumn{2}{c}{Purity} & \multicolumn{2}{c}{Fmeasure}  \\
   \cline{2-9}
Data & SVD & NMF & SVD & NMF & SVD & NMF & SVD & NMF \\
\hline
Reuters2 & $\mathbf{0.495}$ & 0.404 & $\mathbf{0.451}$ & 0.542 & $\mathbf{0.862}$ & 0.822 & $\mathbf{0.860}$ & 0.819 \\ 
Reuters4 & $\mathbf{0.735}$ & 0.629 & $\mathbf{0.349}$ & 0.402 & $\mathbf{0.839}$ & 0.794 & $\mathbf{0.626}$ & 0.562 \\ 
Reuters6 & $\mathbf{0.837}$ & 0.795 & $\mathbf{0.368}$ & 0.384 & 0.682 & $\mathbf{0.745}$ & $\mathbf{0.649}$ & 0.462 \\
Reuters8 & $\mathbf{1.03}$ & 0.923 & $\mathbf{0.319}$ & 0.356 & $\mathbf{0.818}$ & 0.749 & $\mathbf{0.504}$ & 0.404 \\
Reuters10 & $\mathbf{1.18}$ & 1.04 & $\mathbf{0.296}$ & 0.336 & $\mathbf{0.786}$ & 0.731 & $\mathbf{0.516}$ & 0.380 \\
Reuters12 & 1.08 & $\mathbf{1.13}$ & 0.334 & $\mathbf{0.320}$ & 0.684 & $\mathbf{0.739}$ & $\mathbf{0.444}$ & 0.357 \\ 
\hline
  \end{tabular}
  \label{ch2:table5}
 \end{center}
\end{table}

\section{LSI Aspect of the SVD}\label{ch2:lsisvd}

LSI is an indexing method using the truncated SVD to recognize synonyms and polysemes in a document corpus. In this section, LSI aspect of the SVD will be discussed by showing how the SVD handles synonymy and polysemy problems in synthetic datasets. By examining the structure of lower rank approximation of semantic graph's affinity matrix, we will point out that clustering and LSI aspects of the SVD come from the same source: related vertices tend to be more clustered in bipartite graph representation of the approximate matrix than in the original semantic graph. Based on this fact, we will devise an algorithm that has similar functionality as the SVD in clustering the related vertices to handle synonymy and polysemy problems. The proposed algorithm utilizes similarity measures to detect related vertices, and then modifies weights of existing edges and/or creates new edges based on the measures. A numerical analysis will be conducted using standard datasets in LSI reseach to evalute performance of the algorithm.

\subsection{Synonymy} \label{synonym}

Many words in English have the same or almost the same meaning, for example words in \{university, college, institute\}, \{female, girl, woman\}, and \{book, novel, biography\} are synonyms to each other. The classic approach of retrieving relevant documents is by using the \emph{vector space model} \cite{Salton}; a method that transforms a text corpus into a word-by-document matrix where entries are word frequencies in corresponding documents. Since each document vector indexes only words that appear in it, the vector space model cannot retrieve relevant documents containing synonyms of, but not, terms in query. Example in table \ref{ch2:table9} (taken from \cite{Kolda}) describes synonymy problems associated with the vector space model. Note that `Mark Twain' and `Samuel Clemens' refer to the same person, and `purple' and `colour' are closely related. So that reference classes for documents are \{Doc1, Doc2, Doc3\} and \{Doc4, Doc5\}, and for words are \{mark, twain, samuel, clemens\} and \{purple, colour\} with the first/second document class corresponds to the first/second word class respectively.

In the vector space model, task of finding relevant documents to a query is conducted by calculating distances (usually cosine criterion \cite{Berry2}) between query vector $\mathbf{q}\in\mathbb{R}_+^{M\times 1}$ and document vectors $\mathbf{a}_n\in\mathbb{R}_+^{M\times 1}\;\forall n$. The more relevant the document to the query, the closer the distance between them. Query vector is analogous to document vectors; it indexes words that appear both in query and word-by-document matrix $\mathbf{A}=[\mathbf{a}_1,\ldots,\mathbf{a}_N]$. Note that when there are preprocessing steps or the SVD is used to approximate the matrix, then instead of $\mathbf{A}$, approximate matrix $\mathbf{\hat{A}}$ will be used.

\begin{table}
 \begin{center}
   \caption{The vector space model for describing synonymy.}
   \centering
   \begin{tabular}{lrrrrr}
   \hline
Word & Doc1 & Doc2 & Doc3 & Doc4 & Doc5 \\
\hline
mark & 15 & 0 & 0 & 0 & 0 \\ 
twain & 15 & 0 & 20 & 0 & 0 \\ 
samuel & 0 & 10 & 5 & 0 & 0 \\
clemens & 0 & 20 & 10 & 0 & 0 \\
purple & 0 & 0 & 0 & 20 & 10 \\
colour & 0 & 0 & 0 & 15 & 0 \\ 
\hline 
  \end{tabular}
  \label{ch2:table9}
 \end{center}
\end{table}

\begin{table}
 \begin{center}
   \caption{LSI using the SVD for dataset in table \ref{ch2:table9}.}
   \centering
   \begin{tabular}{lrrrrr}
   \hline
Word & Doc1 & Doc2 & Doc3 & Doc4 & Doc5 \\
\hline
mark & 3.72 & 3.50 & 5.45 & 0 & 0 \\ 
twain & 11.0 & 10.3 & 16.1 & 0 & 0 \\ 
samuel & 4.15 & 3.90 & 6.08 & 0 & 0 \\
clemens & 8.30 & 7.80 & 12.2 & 0 & 0 \\
purple & 0 & 0 & 0 & 21.0 & 7.08 \\
colour & 0 & 0 & 0 & 13.5 & 4.55 \\ 
\hline 
  \end{tabular}
  \label{ch2:table10}
 \end{center}
\end{table}

\begin{figure}
 \begin{center}
  \subfigure[]{
   \includegraphics[width=0.45\textwidth]{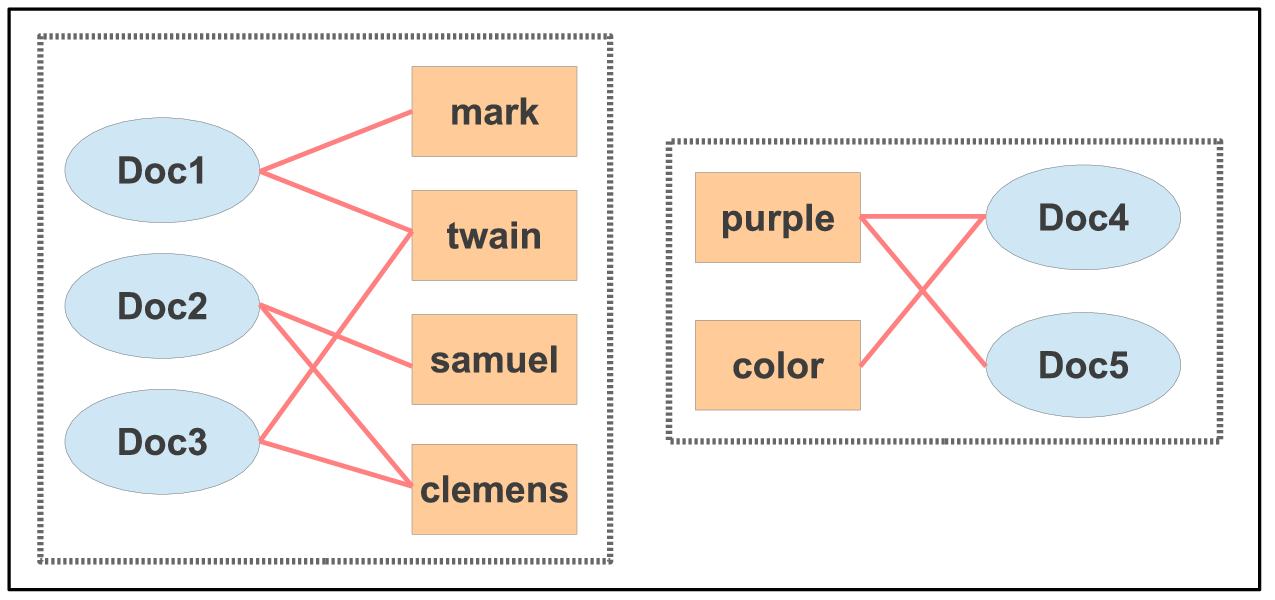}
   \label{synonyma}
  }
  \subfigure[]{
   \includegraphics[width=0.45\textwidth]{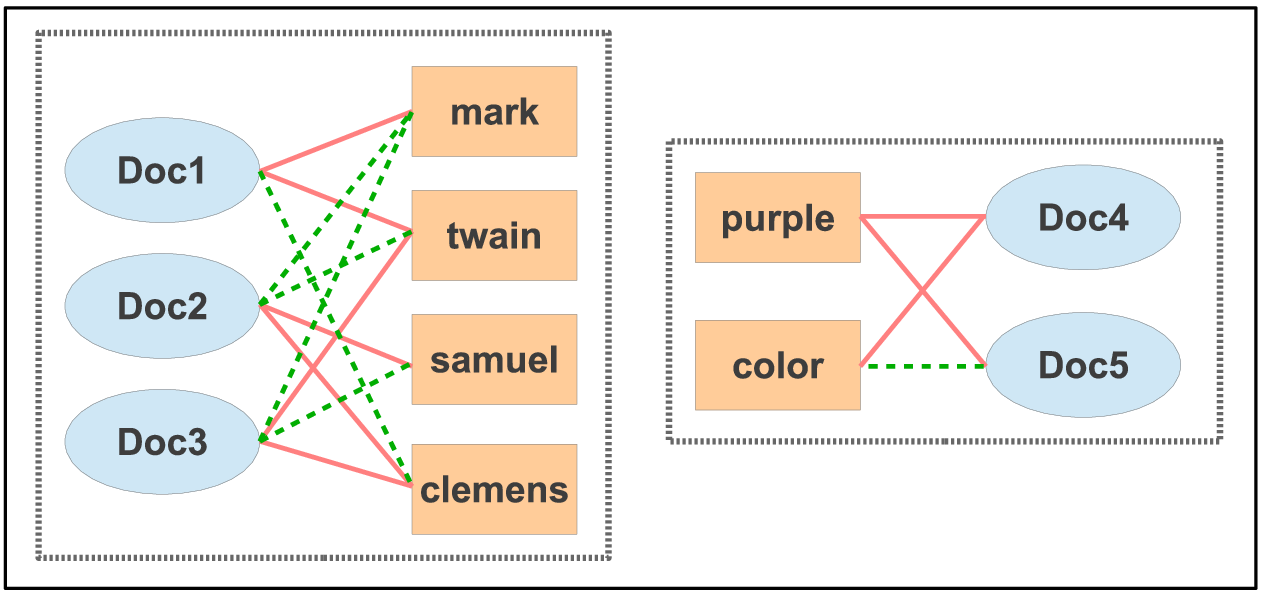}
   \label{synonymb}
  }
  \caption{Bipartite graph representation for (a) the original matrix in table \ref{ch2:table9}, and (b) the approximate matrix in table \ref{ch2:table10}. The dash lines in (b) represent new connections due to lower rank approximation to the original matrix.}
  \label{figsynonym}
 \end{center}
\end{figure}

For the dataset in table \ref{ch2:table9}, when a query containing `mark' and `twain' is created ($\mathbf{q}^T$ $=$ $[1\;$ $1\;$ $0\;$ $0\;$ $0\;$ $0]$), then the result will be $[1.00\;$ $0.00\;$ $0.62\;$ $0.00\;$ $0.00]$ (derived by computing cosine distances between $\mathbf{q}$ and $\mathbf{a}_n\;\,\forall n$). So, only Doc1 and Doc3 will be retrieved; and Doc2 which contains `samuel' and `clemens' (synonyms of `mark' and `twain') won't be recognized as relevant. Similarly, a query containing `colour' but not `purple' won't be able to retrieve Doc4. 

According to a result by Kontostathis and Pottenger \cite{Kontostathis}, LSI using the SVD can recognize synonyms as long as there is a path that chain the synonyms together and the path is close enough. For example in table \ref{ch2:table9} `mark' and `twain' are connected to `samuel' and `clemens' through Doc3. So, there is a short path that connects them together, and thus it can be expected that LSI using the SVD will be able to recognize the synonyms. Similarly, `colour' and `purple' are connected through Doc4, so LSI is also expected to be able to reveal this connection. 

Table \ref{ch2:table10} shows rank-2 SVD approximation of the original matrix in table \ref{ch2:table9} (the rank was chosen based on the number of reference classes). As shown, Doc1, Doc2, and Doc3 are now indexing `mark', `twain', `samuel', and `clemens', and Doc4 and Doc5 are now indexing `purple' and `colour'. Thus, all relevant documents will be correctly retrieved if appropriate queries are made to the system. 

Figure \ref{figsynonym} depicts bipartite graph $\mathcal{G}(\mathbf{A})$ and $\mathcal{G}(\mathbf{\hat{A}})$ representing the original and approximate matrix in table \ref{ch2:table9} and \ref{ch2:table10}. A shown, clusters in $\mathcal{G}(\mathbf{\hat{A}})$ are more connected as the SVD creates new edges between vertices in the same cluster that are not yet connected in the original graph. Hence, from graph model viewpoint, SVD capability in solving synonymy problems can be viewed as its capability in creating new edges between related vertices.

\subsection{Polysemy} \label{polysemy}

Polysemy is the problem of a word with multiple meanings but are not necessarily related. Since a polyseme can appear in unrelated documents, a query containing it will probably also retrieve unrelated documents. Table \ref{ch2:table11} gives an example of such problem where `bank' either refers to financial institution or area near river. By inspection it is clear that reference classes for documents are \{Doc1, Doc3, Doc5\} and \{Doc2, Doc4, Doc6\}, and reference classes for words are \{money, bank, interest\} and \{bed, river, bank\} with the first/second document class corresponds to the first/second word class respectively. 

If query $\mathbf{q}_1$ containing `bank' and `money' (a query corresponds to the first document class) is made to this vector space model, then only Doc1 and Doc3 will be recognized as relevant since the other documents have the same score. Similarly, if query $\mathbf{q}_2$ containing `river' and `bank' (a query corresponds to the second document class) is made, then only Doc2 and Doc4 will be retrieved.

Table \ref{ch2:table12} shows rank-2 SVD approximation of the original matrix in table \ref{ch2:table11}. If $\mathbf{q}_1$ and $\mathbf{q}_2$ are made to the approximate matrix $\mathbf{\hat{A}}$ in table \ref{ch2:table12}, then the result for $\mathbf{q}_1$ will be $[0.77\;$ $0.41\;$ $0.77\;$ $0.41\;$ $0.79\;$ $0.51]$, and for $\mathbf{q}_2$ will be $[0.41\;$ $0.77\;$ $0.41\;$ $0.77\;$ $0.51\;$ $0.79\;]$. So, the SVD is able to handle polysemy problem in this case. And, as now Doc5 indexes `money' and Doc6 indexes `river', all vertices in the same class are connected to each other, and thus any query containing any term in the same word class will be able to retrieve all relevant documents (except when the query contains only the polysemy word `bank').

\begin{table}
 \begin{center}
   \caption{The vector space model for describing polysemy.}
   \centering
   \begin{tabular}{lrrrrrr}
   \hline
Word & Doc1 & Doc2 & Doc3 & Doc4 & Doc5 & Doc6 \\
\hline
money    & 1 & 0 & 1 & 0 & 0 & 0 \\ 
bed      & 0 & 1 & 0 & 1 & 0 & 1 \\ 
river    & 0 & 1 & 0 & 1 & 0 & 0 \\
bank     & 1 & 1 & 1 & 1 & 1 & 1 \\
interest & 1 & 0 & 1 & 0 & 1 & 0 \\
\hline 
  \end{tabular}
  \label{ch2:table11}
 \end{center}
\end{table}

\begin{table}
 \begin{center}
   \caption{LSI using the SVD for dataset in table \ref{ch2:table11}.}
   \centering
   \begin{tabular}{lrrrrrr}
   \hline
Word & Doc1 & Doc2 & Doc3 & Doc4 & Doc5 & Doc6 \\
\hline
money    & $\mathbf{0.809}$ & -0.0550          & $\mathbf{0.809}$ & -0.0550          & $\mathbf{0.547}$ & 0.0621 \\ 
bed      & -0.0239          & $\mathbf{1.08}$ & -0.0239          & $\mathbf{1.08}$ & 0.117            & $\mathbf{0.738}$ \\ 
river    & -0.0550          & $\mathbf{0.809}$ & -0.0550          & $\mathbf{0.809}$ & 0.0621            & $\mathbf{0.547}$ \\
bank     & $\mathbf{1.06}$ & $\mathbf{1.06}$ & $\mathbf{1.06}$ & $\mathbf{1.06}$ & $\mathbf{0.855}$ & $\mathbf{0.855}$ \\
interest & $\mathbf{1.08}$ & -0.0239          & $\mathbf{1.08}$ & -0.0239           & $\mathbf{0.738}$ & 0.117 \\
\hline 
  \end{tabular}
  \label{ch2:table12}
 \end{center}
\end{table}

\begin{figure}
 \begin{center}
  \subfigure[]{
   \includegraphics[width=0.45\textwidth]{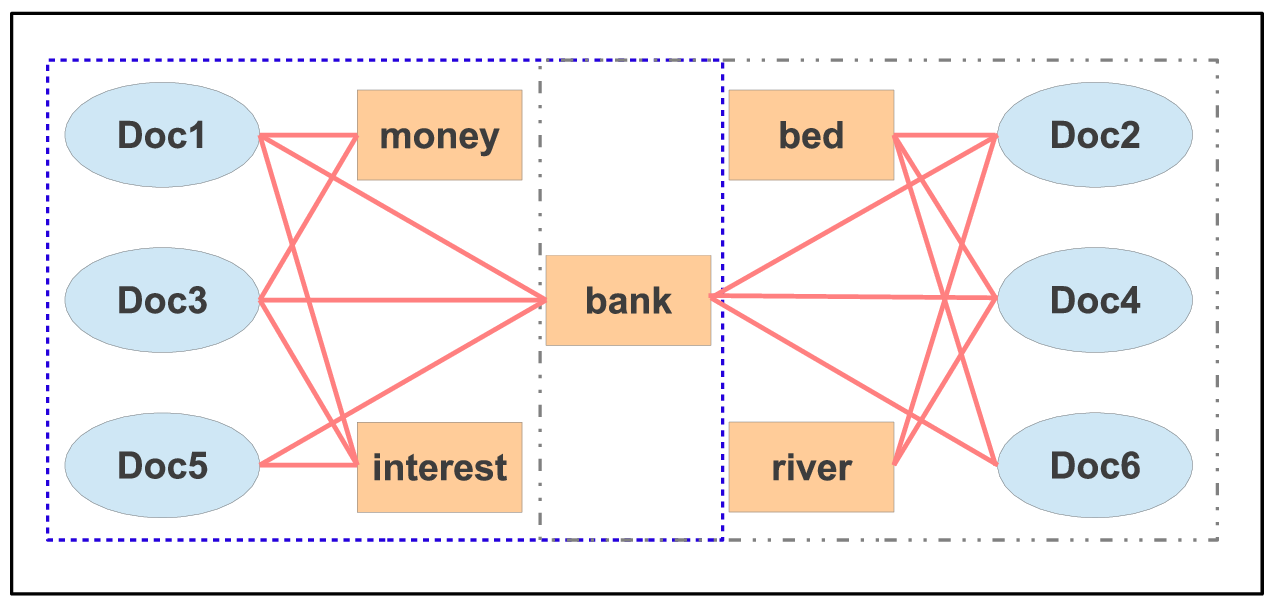}
   \label{polysemya}
  }
  \subfigure[]{
   \includegraphics[width=0.45\textwidth]{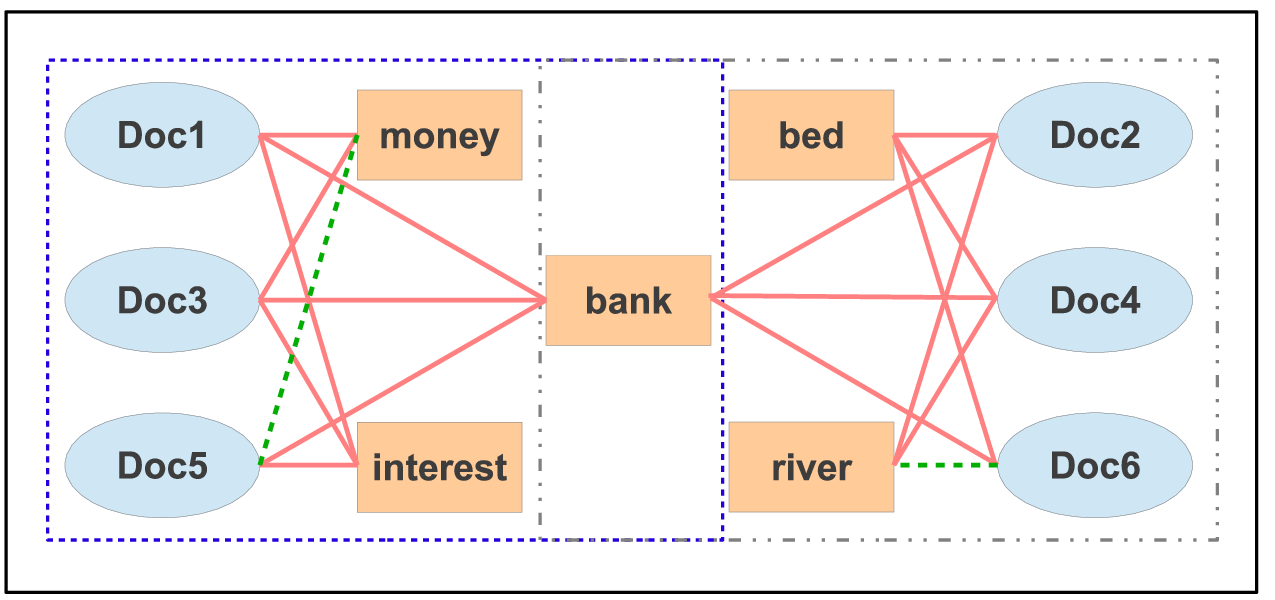}
   \label{polysemyb}
  }
  \caption{Bipartite graph representation for (a) the original matrix in table \ref{ch2:table11}, and (b) the approximate matrix in table \ref{ch2:table12}. The dash lines in (b) represent new connections due to lower rank approximation of the original matrix.}
  \label{figpolysemy}
 \end{center}
\end{figure}

Figure \ref{figpolysemy} shows bipartite graph representations of matrices in table \ref{ch2:table11} and \ref{ch2:table12}. As in the synonymy case, SVD capability in solving the polysemy problem also lies in its capability in recognizing vertices from the same cluster as now there is an edge connecting Doc5 to `money' and Doc6 to `river' respectively. 

From these examples, it can be inferred that LSI aspect of the SVD comes from its capability in grouping similar documents with related words, so that synonyms or related words can be grouped in the same clusters, and the influence of polysemes can be reduced by creating more connected clusters. In the next section, we will devise an algorithm to mimick this capability and demonstrate how the algorithm can also handle the synonymy and polysemy problems.

\subsection{Proposed algorithm}\label{proposedalgorithm}

The proposed algorithm is designed to mimick LSI capability of the SVD by explicitly measuring similarities between word pairs to create more connected clusters. Hence, the algorithm will be more intuitive than the SVD because it determines the connections based on explicit similarity measure. The following describes the algorithm.

Let $\mathbf{A}\in\mathbb{R}_{+}^{M\times N}$ be a word-by-document matrix. By using cosine criterion, similarity between word $p$ and $q$ can be computed by: 
\begin{equation}
s_{pq} = \frac{\mathbf{a}_{p:}\mathbf{a}_{q:}^T}{\|\mathbf{a}_{p:}\|_F\|\mathbf{a}_{q:}\|_F}
\label{similaritymatrix}
\end{equation}
where $\mathbf{a}_{x:}$ denotes $x$-th row of $\mathbf{A}$ and $s_{pq} \in [0, 1]$. If $s_{pq}\rightarrow 1$, then $p$ and $q$ are strongly related since they coappear in many documents, and if $s_{pq}\rightarrow 0$ then $p$ and $q$ are unrelated.

The proposed algorithm works by propagating entry weights of word vectors to each other based on similarity measures between the vectors. We define the following update rule to update entry $j$ of word $i$:
\begin{equation}
a_{ij} \leftarrow \max(a_{ij}, s_{ik}a_{kj})\;\,\forall k\ne i.
\label{updaterule}
\end{equation}
As shown, if word $i$ and $k$ are related, i.e., $s_{ik}>0$, then the rule can make word $i$ and $k$ co-appear in documents that index either $i$ or $k$. Consequently, it can produce more clustered graph as in the SVD case.

We use maximum rather than average value in the update rule. Certainly, this mechanism will not be able to scale the entries down to reduce influence of high frequency words which usually is preferable in document processing. There are several reasons that make maximum value is more advantageous in this case:
\begin{itemize}
\item High frequency words usually are the stop words (words that have little influence in conveying document contents) that will be removed in the preprocessing step.
\item Logarithmic scale usually is used to balance the entry weights so that influence of high frequency words can be reduced.
\item As will be shown later,  the convergence of the algorithm can be established if maximum value is used to update the entries.
\end{itemize}

The following outlines the algorithm, where $\mathbf{A}^{(0)}$ denotes the initial matrix, $a_{ij}^{(0)}$ denotes $(i,j)$ entry of $\mathbf{A}^{(0)}$, $a_{ij}^{(n)}$ denotes $a_{ij}$ value at $n$-th iteration, and $maxiter$ denotes maximum number of iteration. Because the algorithm replaces some zero entries with positive numbers as the update process progresses, we name it as `similarity-based matrix completion algorithm'.

\begin{algorithm}[H]
\caption{Similarity-based matrix completion algorithm.}
\label{matrixcompletionalgorithm}
\begin{algorithmic}
\STATE \begin{enumerate}
\item Input: $\mathbf{A}^{(0)}\in\mathbb{R}^{M\times N}_{+}$.
\item Construct word similarity matrix $\mathbf{S}\in\mathbb{R}^{M\times M}_{+}$ which entry $s_{pq}$ is computed using eq.~\ref{similaritymatrix}.
\item Update entry $a_{ij}$ using the following procedure:\\
\bf{for} $n = 1,\ldots,maxiter$ \bf{do}\\
$\;\;\;\;a_{ij}^{(n)} \leftarrow \max(a_{ij}^{(n-1)}, s_{ik}a_{kj}^{(n-1)}),\;\,\forall i,j,k\ne i$.\\
\bf{end for}
\end{enumerate}
\end{algorithmic}
\end{algorithm}

The following theorem states convergence property of the algorithm.
\begin{theorem} \label{theorem5}
Given that $\mathbf{A}^{(0)}$ is a nonnegative bounded matrix and there are sufficiently small number of word pairs that are perfectly similar (i.e., $s_{pq} = 1$). Then for each $(i,j)$, sequence $\{a_{ij}^{(n)}\}$ generated by algorithm \ref{matrixcompletionalgorithm} converges to a unique nonnegative bounded value within finite number of iteration.
\end{theorem}
\begin{proof}
Since the update process only involves entries from the same column, it is sufficient to analyze any column of $\mathbf{A}$. Consider that $\mathbf{a}$ = $[a_1,$ $\ldots$ $,a_M]$ to be a column of $\mathbf{A}$, then entry $a_i$ will be updated by the following steps:
\begin{align*}
a_i^{(1)} &= \max(a_{i}^{(0)}, s_{ik}a_{k}^{(0)})\;\,\forall k\ne i\\
a_i^{(2)} &= \max(a_{i}^{(1)}, s_{ik}a_{k}^{(1)})\;\,\forall k\ne i\\
&\vdots\\
a_i^{(n)} &= \max(a_{i}^{(n-1)}, s_{ik}a_{k}^{(n-1)})\;\,\forall k\ne i.\\
\end{align*}

Since $\mathbf{A}^{(0)}$ is a nonnegative bounded matrix, then $\mathbf{A}^{(n)}\,\forall n$ will also be a nonnegative bounded matrix. Thus, we need only to prove convergence and uniqueness of $a_i^{(n)}\,\forall i$ for sufficiently large $maxiter$. 

The following describes step-by-step update process for $a_i$. First iteration: 
\begin{align*}
a_i^{(1)} = \max(a_{i}^{(0)}, s_{ik}a_{k}^{(0)})\;\,\forall k\ne i.
\end{align*}
Second iteration: $a_i^{(2)}$ = $\max(a_{i}^{(1)}$, $s_{ik}a_{k}^{(1)})\;\,\forall k\ne i$, where $a_k^{(1)}$ = $\max(a_{k}^{(0)}$, $s_{kj}a_{j}^{(0)})\;\,\forall j\ne k$, so that 
\begin{align*}
a_i^{(2)} = \max(a_{i}^{(1)}, s_{ik}s_{kj}a_{j}^{(0)})\;\,\forall k\ne i, \forall j\ne k.
\end{align*}
Third iteration: $a_i^{(3)}$ = $\max(a_{i}^{(2)}$, $s_{ik}a_{k}^{(2)})\;\,\forall k\ne i$, where $a_k^{(2)}$ = $\max(a_{k}^{(1)}$, $s_{kj}a_{j}^{(1)})\;\,\forall j\ne k$, and $a_j^{(1)}$ = $\max(a_{j}^{(0)}$, $s_{jl}a_{l}^{(0)})\;\,\forall l\ne j$. So that:
\begin{align*}
a_i^{(3)} = \max(a_{i}^{(2)}, s_{ik}s_{kj}s_{jl}a_{l}^{(0)})\;\,\forall k\ne i,\;\,\forall j\ne k,\;\,\forall l\ne j .
\end{align*}
Thus, in general $a_{i_1}$ at $n$-th iteration can be written as:
\begin{align}
a_{i_1}^{(n)} = \max(a_{i_1}^{(n-1)}, s_{i_1i_2}s_{i_2i_3}\cdots s_{i_{n-1}i_{n}}a_{i_{n}}^{(0)})\;\,\forall i_2\ne i_1, \forall i_3\ne i_2, \ldots, \forall i_{n}\ne i_{n-1}.
\label{updaterulen}
\end{align}

As shown, if there are sufficiently small number of word pairs with perfect similarity, then $a_{i_1}^{(n-1)}$ $\ge$ $s_{i_1i_2}s_{i_2i_3}\cdots s_{i_{n-1}i_{n}}a_{i_{n}}^{(0)}$ can always be guaranteed for sufficiently large $n$. And consequently the convergence of the algorithm is guaranteed.

Further because in each iteration $a_i^{(n)}$ is unique (either $a_i^{(n-1)}$ or $s_{ik}a_{k}^{(n-1)}$), then the uniqueness of the solution can always be guaranteed.
\end{proof}

Notice that since $a_{ij}^{(n)}\ge a_{ij}^{(n-1)}\;\,\forall i, j, n$, then it is possible to have $a_{ij}^{(n)} = a_{ij}^{(n-1)}\;\,\forall i, j,\exists n$ before convergence. Thus, it is necessary to run the update procedure a few times after $\mathbf{A}^{(n)} = \mathbf{A}^{(n-1)}$ condition has been achieved to avoid false convergence.

\begin{figure}
 \begin{center}
  \includegraphics[width=1.0\textwidth]{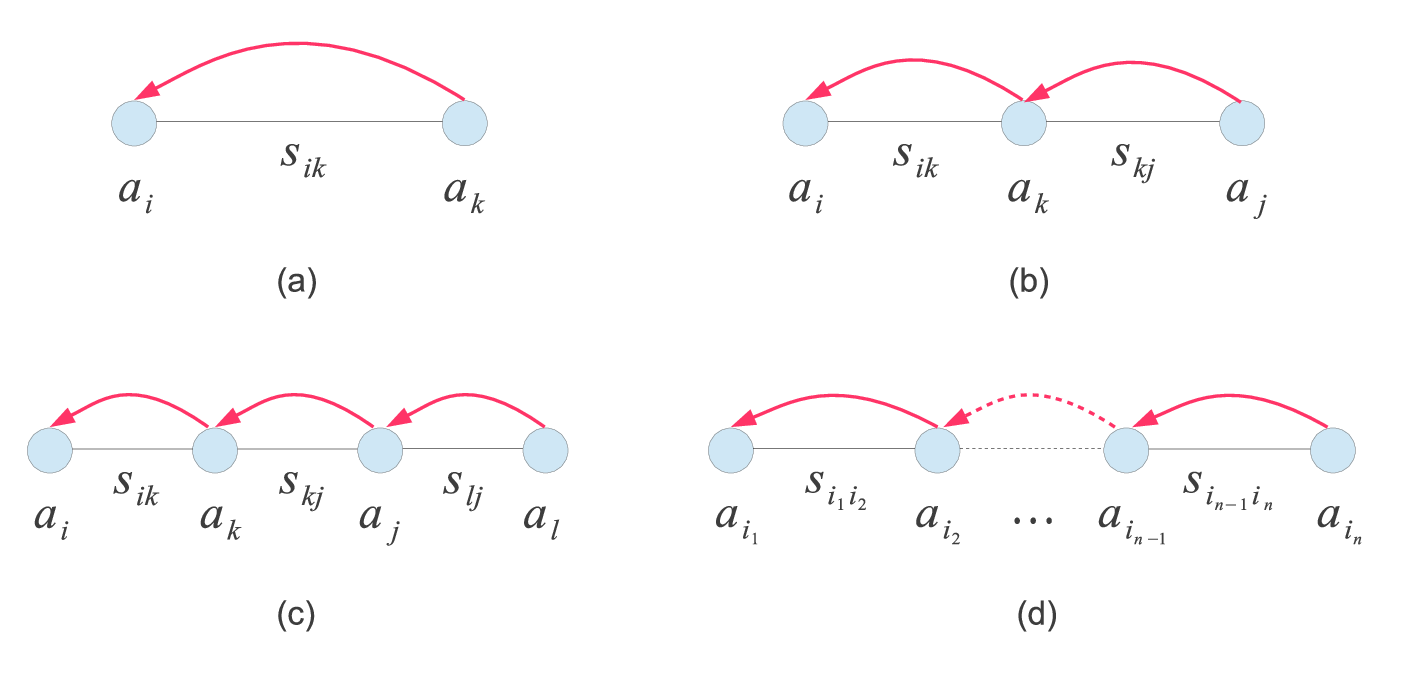}
  \caption{Weight propagation process, (a) in the first iteration, (b) in the second iteration, (c) in the third iteration, and (d) in the $n$-th iteration.}
  \label{weightpropagation}
 \end{center}
\end{figure}

Update procedure in $n$-th iteration (eq.~\ref{updaterulen}) reveals the nature of weight propagation process from $a_{i_{n}}$ to its most similar entry $a_{i_{n-1}}$ until reaching the current entry $a_{i_1}$. This process can be thought as a capability of sensing synonym chain in the document collection which is the reason why we do not need to consider more than one synonym of each word in the update rule. Figure \ref{weightpropagation} depicts the weight propagation process as described by eq.~\ref{updaterulen}.

Computational complexity of the algorithm is the sum of complexity for computing $\mathbf{S}$, i.e., $O(M\times M\times N)$, and complexity for running the update procedure until convergence, i.e., $O(maxiter\times M\times M\times N)$. However, as will be shown in the next section, in real datasets usually percentages of word pairs with perfect similarity are very small so that the algorithm converges in a few iterations, and thus the average computational complexity of the algorithm is $\Theta(M\times M\times N)$. In addition, the algorithm can be sped up by only executing the update procedure if $s_{ik} \ne 0$. Table \ref{completionsynonymy} and \ref{completionpolysemy} show results of utilizing the algorithm to solve the synonymy and polysemy problems described previously. As shown, the algorithm also can handle the problems.

\begin{table}
 \begin{center}
   \caption{LSI using algorithm \ref{matrixcompletionalgorithm} for synonymy problem.}
   \centering
   \begin{tabular}{lrrrrr}
   \hline
Word & Doc1 & Doc2 & Doc3 & Doc4 & Doc5 \\
\hline
mark & 15 & 4.3 & 12 & 0 & 0 \\ 
twain & 15 & 7.2 & 20 & 0 & 0 \\ 
samuel & 5.4 & 20 & 10 & 0 & 0 \\
clemens & 5.4 & 20 & 10 & 0 & 0 \\
purple & 0 & 0 & 0 & 20 & 10 \\
colour & 0 & 0 & 0 & 18 & 8.9 \\ 
\hline 
  \end{tabular}
  \label{completionsynonymy}
 \end{center}
\end{table}

\begin{table}
 \begin{center}
   \caption{LSI using algorithm \ref{matrixcompletionalgorithm} for polysemy problem.}
   \centering
   \begin{tabular}{lrrrrrr}
   \hline
Word & Doc1 & Doc2 & Doc3 & Doc4 & Doc5 & Doc6 \\
\hline
money    & \bf{1} & 0.58   & \bf{1} & 0.58   & \bf{0.82} & 0.58      \\ 
bed      & 0.71   & \bf{1} & 0.71   & \bf{1} & 0.71      & \bf{1}    \\ 
river    & 0.58   & \bf{1} & 0.58   & \bf{1} & 0.58      & \bf{0.82} \\
bank     & \bf{1} & \bf{1} & \bf{1} & \bf{1} & \bf{1}    & \bf{1}    \\
interest & \bf{1} & 0.71   & \bf{1} & 0.71   & \bf{1}    & 0.71      \\
\hline 
  \end{tabular}
  \label{completionpolysemy}
 \end{center}
\end{table}

\subsection{Experimental results}\label{ch2:lsiresults}

We will now evaluate LSI capability of the proposed algorithm using standard datasets in LSI research\protect\footnote{The datasets can be downloaded at http://web.eecs.utk.edu/research/lsi/}. Table \ref{ch2:table16} summarizes some information of the datasets, where \#Documents, \#Words, \%NNZ, \%PS, and \#Queries respectively denote the number of documents, the number of unique words, percentage of nonzero entries in corresponding word-by-document matrix, percentage of word pairs with perfect similarity, and the number of predefined queries made to corresponding word-by-document matrix. Note that the experiments were conducted using the same hardware and platform as in section \ref{ch2:results1}.

\begin{table}
 \begin{center}
   \caption{Some information of the datasets.}
   \centering
   \begin{tabular}{lrrrr}
   \hline
            & Medline & Cranfield & CISI    & ADI    \\
\hline
\#Documents & 1033   & 1398   & 1460   & 82    \\ 
\#Words     & 12011  & 6551   & 9080   & 1215  \\ 
\%NNZ       & 0.457  & 0.857  & 0.517  & 2.15  \\
\%PS        & 0.0434 & 0.0244 & 0.0289 & 0.631 \\
\#Queries   & 30     & 225    & 35     & 35   \\
\hline 
  \end{tabular}
  \label{ch2:table16}
 \end{center}
\end{table}

Each of the text collections in table \ref{ch2:table16} comprises of three important files. The first file contains abstracts of the documents which each is indexed by a unique identifier. The second file contains a list of queries which each has a unique identifier. And the third file contains a dictionary that maps every query to its manually assigned relevant documents. 

The first file is the file that was used to construct word-by-document matrix $\mathbf{A}\in\mathbb{R}_+^{M\times N}$. To extract unique words, stop words\protect\footnote{http://snowball.tartarus.org/algorithms/english/stop.txt} and words that shorter than two characters were removed. But we did not employ any stemming and did not remove words that only belong to one documents as in section \ref{ch2:results1}. The reasons are stemming seems to be not popular in LSI research, and removing unique words can reduce the recall since queries can contain the words. After $\mathbf{A}$ was constructed, we further adjusted the entry weights by using logarithmic scale, i.e., $A_{ij}\leftarrow \log(A_{ij}+1)$, but did not normalized columns of the matrix. This is because based on our pre-experimental results, logarithmic scale improved the retrieval performance, and normalization reduced the retrieval performance for both the SVD and the proposed algorithm in all text collections.

The second file is the file that was used to construct query matrix $\mathbf{Q}$ $\in$ $\mathbb{R}_+^{Q\times M}$ $=$ $[\mathbf{q}_1,$ $\ldots,$ $\mathbf{q}_Q$ $]^T$ where $Q$ denotes the number of queries (shown in the last row of table \ref{ch2:table16}), $M$ denotes the number of unique words which is the same with the number of unique words in corresponding $\mathbf{A}$, and $\mathbf{q}_q$ ($q\in[1, Q]$) denotes $q$-th query vector constructed from the file. Thus, one can get a matrix that contains scores that describe relevancy between documents to a query in corresponding row by multiplying $\mathbf{Q}$ with $\mathbf{A}$.

And the information in the third file was utilized as references to measure retrieval performance quality.

Recall and precision are the most commonly used metrics to measure IR performance. Recall measures proportion of retrieved relevant documents so far to all relevant documents in the collection. And precision measures proportion of retrieved relevant documents to all retrieved documents so far. Recall is usually not indicative of retrieval performance since it is trivial to get perfect recall by retrieving all documents. And as discussed by Kolda and O'Leary \cite{Kolda}, \emph{pseudo-precision} at predefined recall level captures both recall and precision concepts. We used a modified version of this metric known as \emph{average precision}\cite{Harman}, a standard metric in IR research that measures $I$-point interpolated average pseudo-precision at recall level $[0,1]$. In the following, the definition and formulation of the metric is outlined. Detailed discussion can be found in, e.g., ref.~\cite{Harman, Kolda, Berry3}.

Let $\mathbf{r}=\mathbf{q}^T\mathbf{A}$ be sorted in descending order. The precision at $n$-th document is given by:
\begin{equation*}
p_n = \frac{r_n}{n}.
\end{equation*}
where $r_n$ denotes the number of relevant documents up to $n$-th position. The pseudo-precision at recall level $x\in[0,1]$ is defined as:
\begin{equation*}
\hat{p}(x) = \max\{p_n\;|\;x\le r_n/r_N,\;\,n=1,\ldots,N\},
\end{equation*}
where $r_N$ denotes the total number of relevant documents in the collection. The $I$-interpolated average pseudo-precision at recall level $x\in[0,1]$ for a single query is defined as:
\begin{equation*}
\frac{1}{I}\sum_{n=0}^{I-1}\hat{p}\left(\frac{n}{I-1}\right),
\end{equation*}
where as stated previously, $n$ denotes the $n$-th position in $\mathbf{r}$. We used 11-point interpolated average precision as proposed in ref.~\cite{Kolda} because three out of four datasets used here are similar to those used in ref.~\cite{Kolda}.

\begin{figure}
 \begin{center}
  \includegraphics[width=0.7\textwidth]{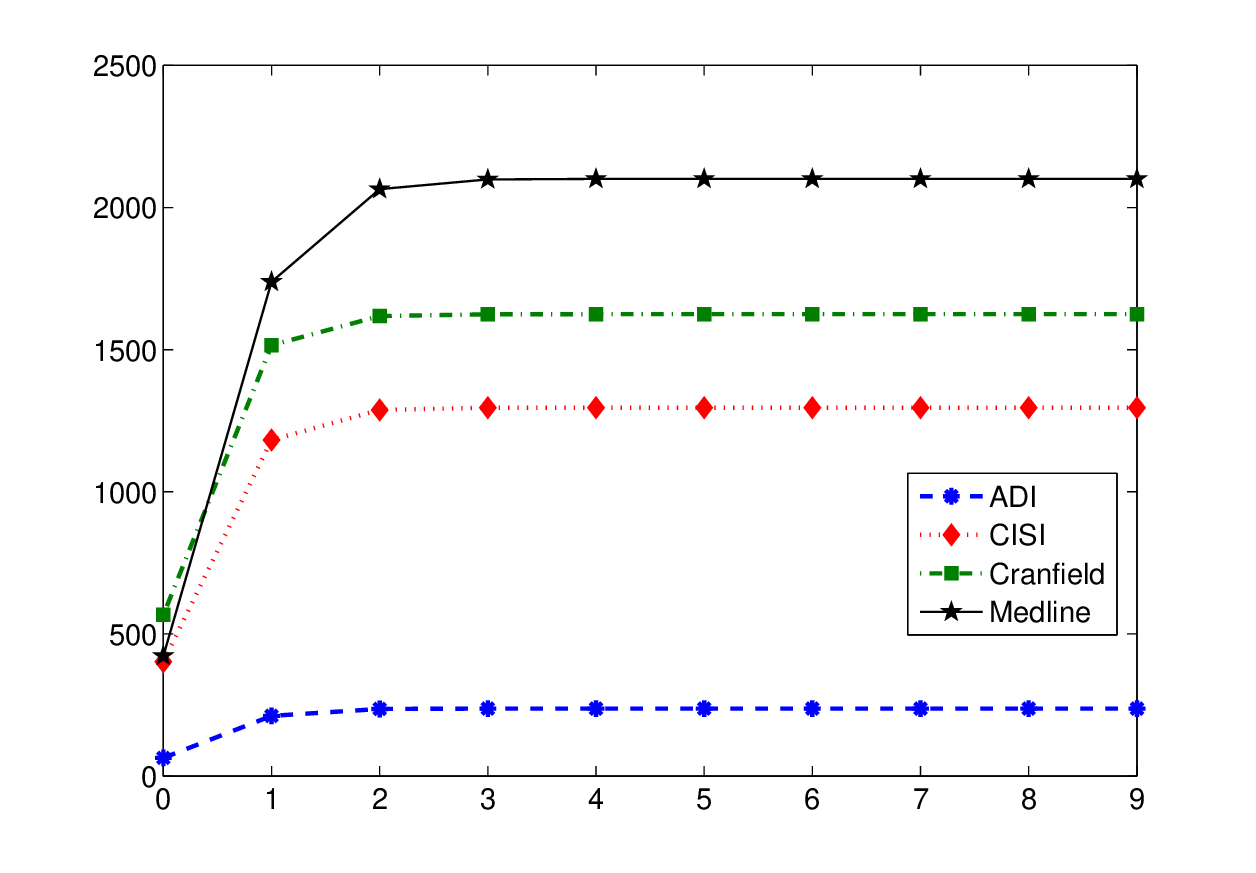}
  \caption{Frobenius norms per iteration as the update procedure progresses.}
  \label{matrixnorms}
 \end{center}
\end{figure}

Because every entry of matrix $\mathbf{A}$ is monotonically nondecreasing, $\|\mathbf{A}^{(n)}\|_F$ = $\|\mathbf{A}^{(n+1)}\|_F$ condition can be used as a starting point to check whether the convergence has been reached. Figure \ref{matrixnorms} shows Frobenius norms of $\mathbf{A}^{(n)}\;\,\forall n=[0, 9]$ for all datasets. As shown, the norm values have been converged at 6-th, 5-th, 4-th, and 5-th iteration for Medline, Cranfield, CISI, and ADI respectively. However as noted previously, it is still necessary to run the update procedure for some more iterations to avoid false convergence. To determine the number of iterations at which the algorithm has converged ($conviter$), let $u$ be the iteration when the norm has converged. Then $conviter\ge u$ is the first iteration when $a_{ij}^{(n)} = a_{ij}^{(n+1)} = \cdots = a_{ij}^{(n+m)}\;\,\forall i,j$ for some $n\ge u$ and $m$. 

Table \ref{proposedalgstatistics} displays the number of iterations and computational times (in seconds) at the convergence for every dataset. As shown, the algorithm needed only a few number of iterations to reach convergence due to the percentage of perfect similarities is very small in every dataset. And because the complexity of the algorithm is cubic, there are significant leaps in the computational times from the smallest dataset, ADI, to the other datasets (note that the sizes of Medline, Cranfield, and CISI matrices are about 125, 92, and 133 times of the size of ADI matrix).

\begin{table}
 \begin{center}
   \caption{The number of iterations and computational times (in seconds) of algorithm \ref{matrixcompletionalgorithm} at the convergence.}
   \centering
   \begin{tabular}{lrrrr}
   \hline
                      & Medline & Cranfield & CISI  & ADI   \\
\hline
\#iteration  & 6       & 5         & 4     & 5    \\
Computational times   & 2015    & 1091      & 1435  & 1.99  \\
\hline 
  \end{tabular}
  \label{proposedalgstatistics}
 \end{center}
\end{table}

Computational times for the SVD were not calculated for each decomposition rank. Rather we computed full rank SVD using MATLAB built-in SVD function for each dataset and recorded the times which were 499, 449, 237, and 1.19 seconds for Medline, Cranfield, CISI, and ADI respectively (these values suggest that the SVD function also has a cubic complexity), and then constructed rank-$k$ truncated SVD by taking the first $k$ columns of the singular matrices and $k\times k$ principal submatrix of the singular value matrix. This approach, thus, is much more efficient because we need many truncated SVDs for each dataset to find the best approximate matrix in term of retrieval performance. 

In summary, the proposed algorithm is about three and half times slower than the SVD in average. These results, however, are not conclusive for two reasons. The first is because we used MATLAB built-in SVD function which is highly optimized, and wrote an unoptimized implementation of our algorithm. And more importantly as noted above, the second is retrieval performances of the SVD must be evaluated over many decomposition ranks to obtain acceptable results (here for Medline, Cranfield, and CISI the ranks were $k\in[10,20,\ldots,600]$, and for ADI the ranks were $k\in[1,2,\ldots,40]$). Depending on the number of the ranks (which can be as large as the number of rows or columns of the input matrix itself), the SVD can be slower than the proposed algorithm.

Table \ref{ch2:table17} shows 11-point interpolated average precision values comparison between the SVD and the proposed algorithm with the values displayed for the SVD are in format \emph{bestval} (\emph{rank}), where \emph{bestval} denotes the best value over all the ranks and \emph{rank} is the rank at \emph{bestval}. Because there are several queries for each text collection (shown in the last row in table \ref{ch2:table16}), the displayed values are the average values over these queries. As shown, the performances of the proposed algorithm and the SVD are comparable. 

\begin{table}
 \begin{center}
   \caption{11-point interpolated average precision.}
   \centering
   \begin{tabular}{lllll}
   \hline
             & Medline & Cranfield & CISI & ADI    \\
\hline
SVD & 0.4967 (600) & 0.3365 (600) & 0.1617 (170) & 0.2663 (33) \\
Algorithm \ref{matrixcompletionalgorithm} & 0.4888 & 0.3537 & 0.1559 & 0.3185 \\
\hline 
  \end{tabular}
  \label{ch2:table17}
 \end{center}
\end{table}

\begin{figure}
 \begin{center}
  \subfigure[Medline]{
   \includegraphics[width=0.45\textwidth]{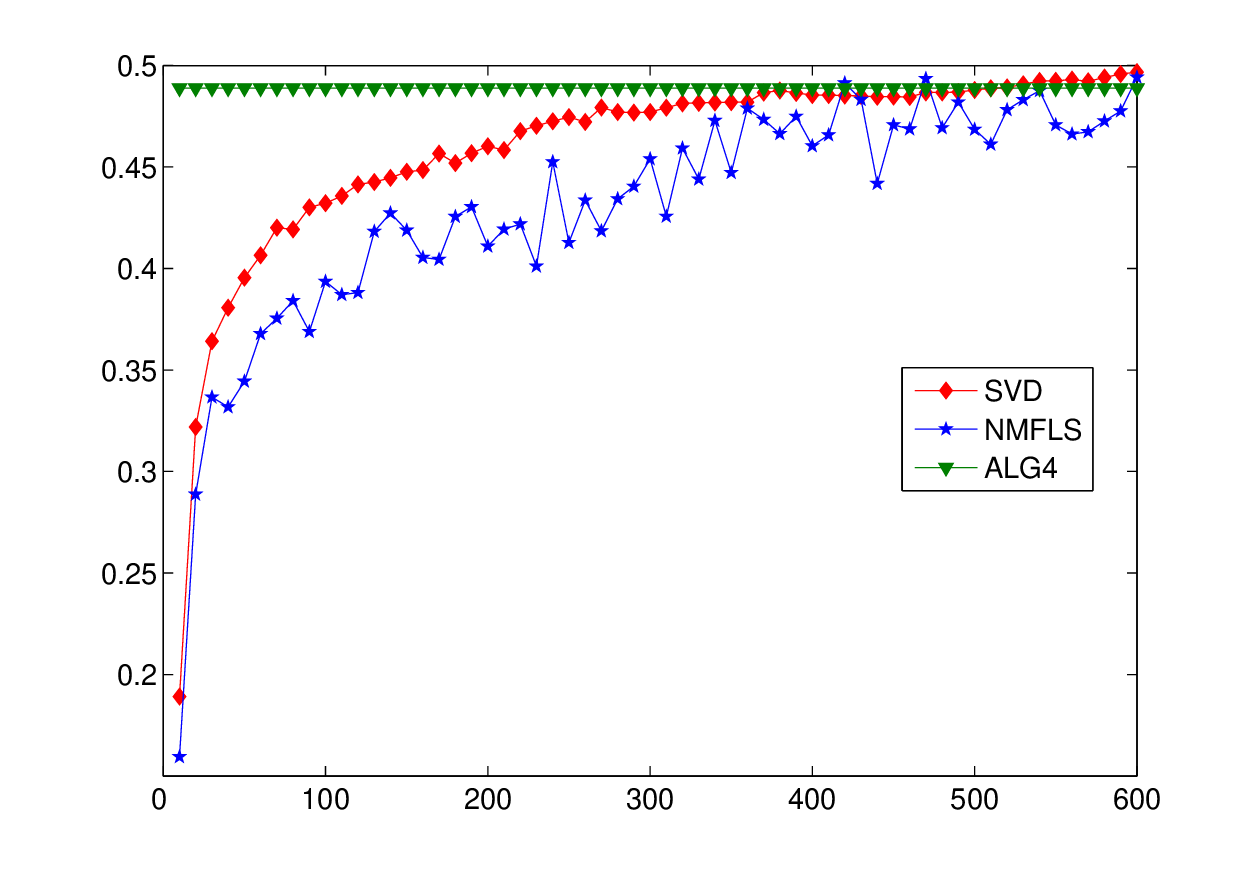}
   \label{ch3:fig3a}
  }
  \subfigure[Cranfield]{
   \includegraphics[width=0.45\textwidth]{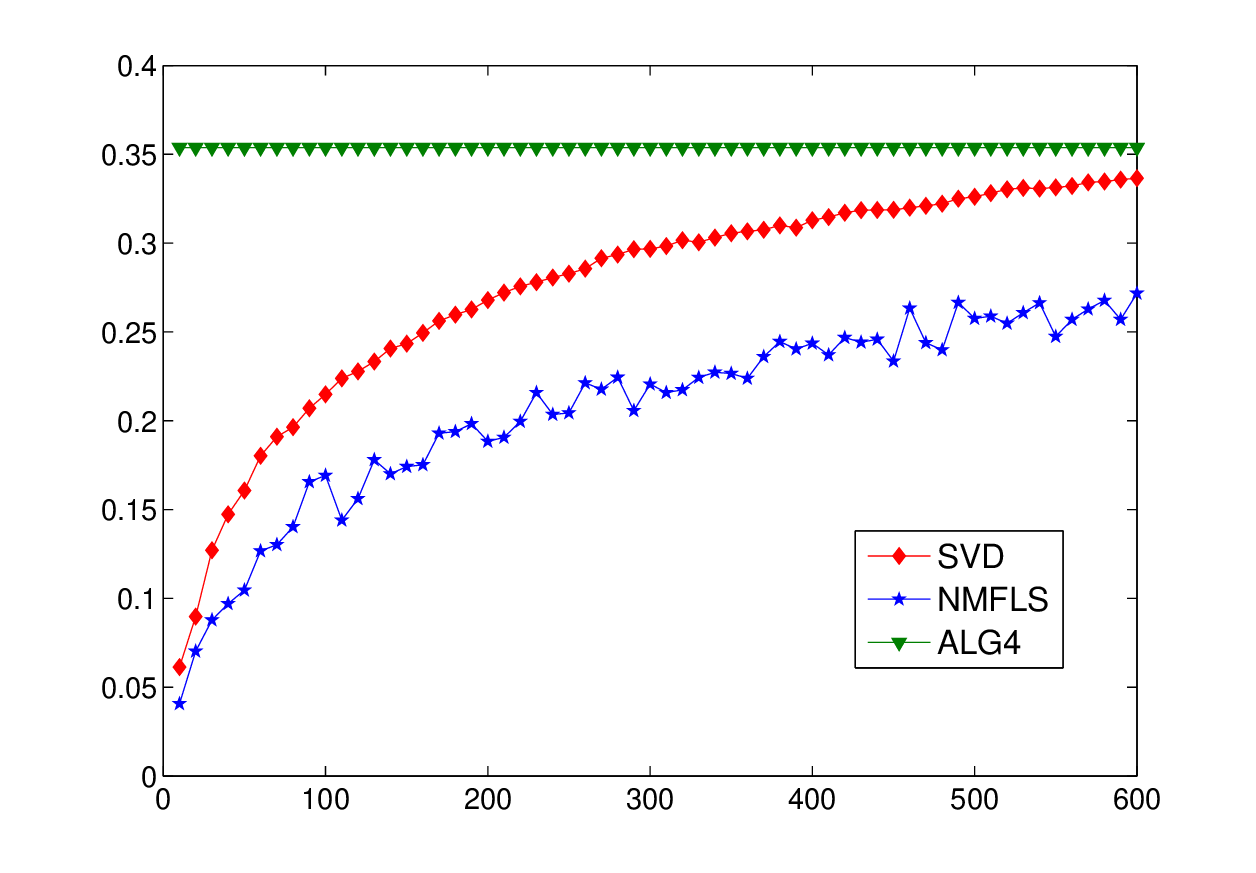}
   \label{ch3:fig3b}
  }
\\
  \subfigure[CISI]{
   \includegraphics[width=0.45\textwidth]{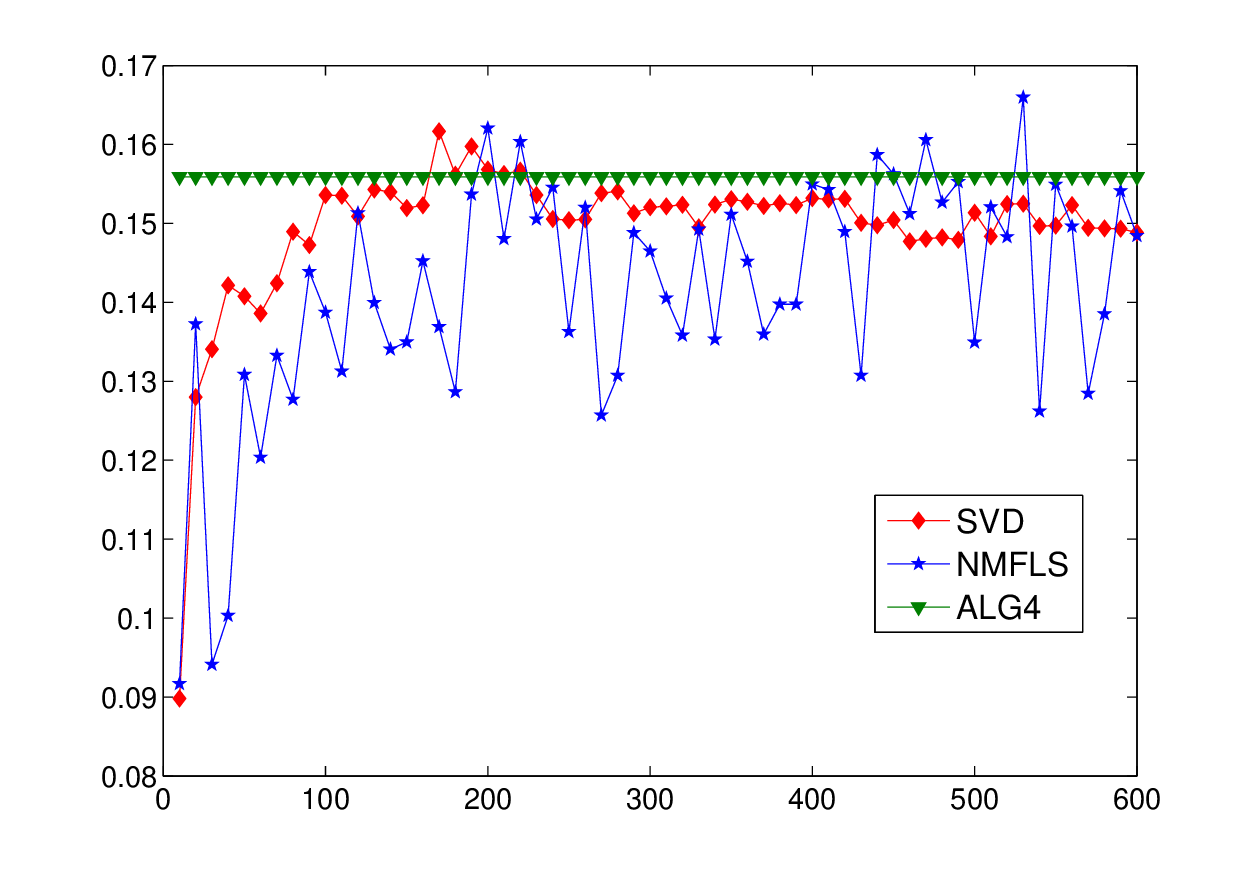}
   \label{ch3:fig3c}
  }
  \subfigure[ADI]{
   \includegraphics[width=0.45\textwidth]{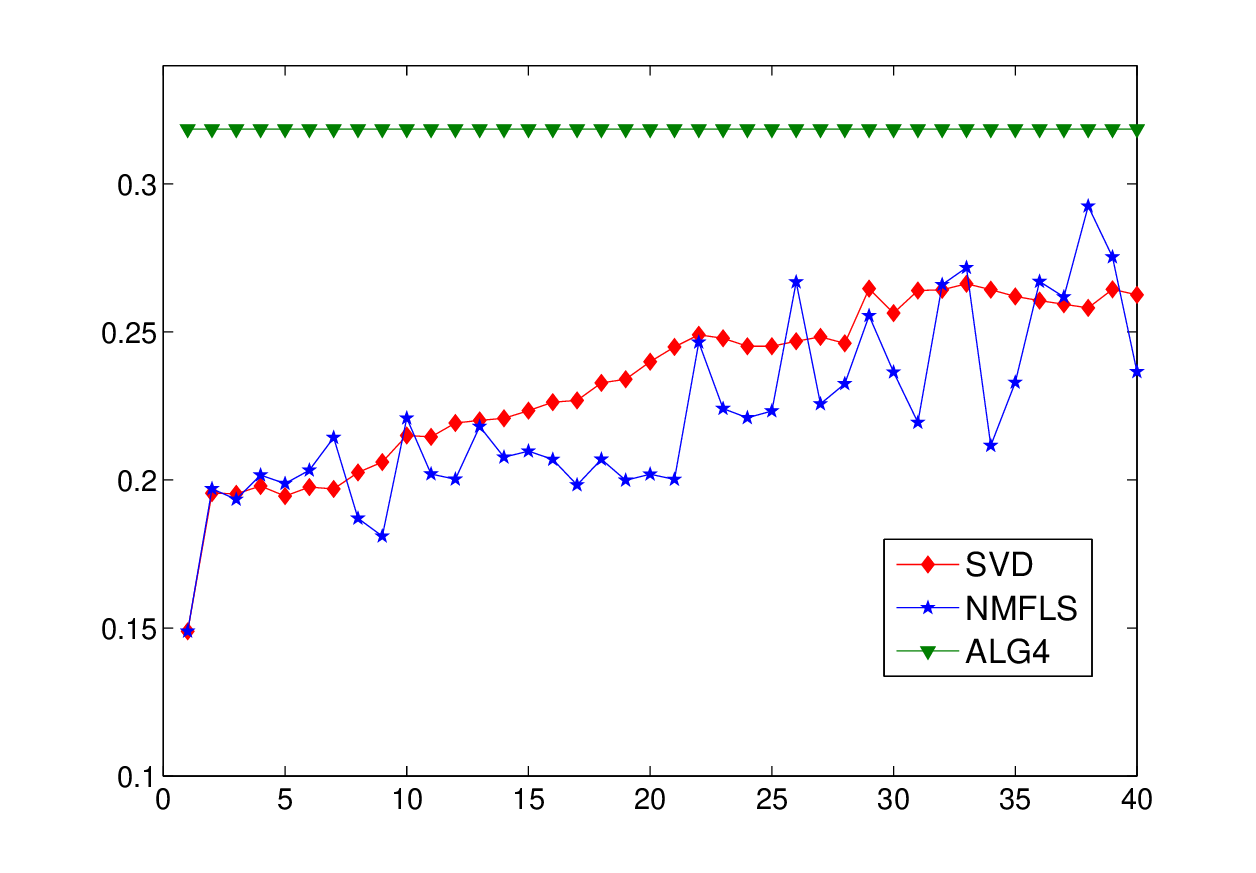}
   \label{ch3:fig3d}
  }
  \caption{11-point interpolated average precision values over decomposition ranks where ALG4 denotes the proposed algorithm and NMFLS denotes NMF algorithm by Lee and Seung \cite{Lee2}.}
  \label{averageprecision}
 \end{center}
\end{figure}

Finally, figure \ref{averageprecision} shows 11-point interpolated average precision values over decomposition ranks for the SVD. Since the proposed algorithm is not a matrix decomposition technique, we appended the NMF into the plots for comparison purpose. As discussed previously, the NMF is similar to the SVD in term that it is a matrix decomposition technique that can be used for clustering. So it can be expected that the NMF also has LSI capability. As shown, in general the SVD outperformed the NMF. But both the SVD and NMF can only outperformed the proposed algorithm when optimal decomposition ranks were used to construct the approximate matrices.

\section{Conclusions}

We have discussed clustering and LSI aspects of the SVD. In clustering part, we suggested that the Ky Fan theorem is the theoretical basis for the using of the SVD in clustering unipartite graph or other graphs with corresponding affinity matrices have been transformed into symmetric matrices. By extending this theorem to complex rectangular matrix, we built a theoretical support for the direct use of the SVD in clustering bipartite and directed graph. Experimental results using synthetic and real datasets confirmed SVD capability in clustering graphs with symmetric and rectangular affinity matrices as supported by the theorem and its extension.

In LSI part, by analyzing the structure of lower rank approximate matrices, we showed that LSI aspect of the SVD is originated from its capability in strengthening connections within clusters. Accordingly, both clustering and LSI aspects of the SVD actually come from the same source, i.e., its clustering capability. This inspired us to develop a clustering-based LSI algorithm.

The proposed algorithm uses cosine criterion to measure degree of similarity between vertex pairs, and then adjusts edge weights and/or creates new edges based on the similarity measures. Convergence analysis showed that the algorithm is convergent and produces a unique result for each input matrix. Computational complexity of the algorithm is cubic, i.e., $\Theta(M\times M\times N)$ with $M\times N$ denotes the size of input matrix. Numerical evaluation using synthetic datasets showed that the algorithm can also recognize synonyms and polysemes. Experimental results using standard datasets in LSI research showed that the algorithm has comparable retrieval performances but was about three and half times slower than the SVD. However, since there is no need to evaluate retrieval performances over some decomposition ranks, depending on the number of the ranks and the size of the datasets, the algorithm can be faster than the SVD.

\section*{Acknowledgement}
This research was partially supported by  Universiti Teknologi Malaysia under Research University Grant number R.J130000.7728.4D064.





\bibliographystyle{plain}
\bibliography{paper}






\end{document}